\pdfoutput=1

\documentclass[11pt]{article}
\usepackage[usenames,dvipsnames]{color} 
\usepackage[table]{xcolor}
\usepackage{tabularx}

\usepackage[final]{acl}

\usepackage{times}
\usepackage{latexsym}
\usepackage{amssymb}
\usepackage{pgfplots}
\usepackage{scalefnt}
\usepackage{boldline}
\usepackage{multirow}
\usepackage{makecell}
\usepackage{graphicx}
\usepackage{bm}
\usepackage{amsmath}
\usepackage{bbm}
\usepackage{float}
\usepackage{stfloats}
\usepackage{lipsum}
\usepackage{comment}
\usepackage{booktabs}
\usepackage{subfig}
\usepackage{lipsum}
\usepackage{enumitem}
\usepackage{soul}
\usepackage[most]{tcolorbox}
\usepackage{colortbl}
\usepackage[flushleft]{threeparttable}

\usepackage{float}
\usepackage{hyperref}       
\usepackage{url}            

\usepackage[T1]{fontenc}

\usepackage[utf8]{inputenc}

\usepackage{microtype}

\usepackage{inconsolata}

\usepackage{graphicx}

\usepackage{soul}

\newcommand{\Fwd}{Forward discrepancy}
\newcommand{\Bwd}{Backward discrepancy}
\newcommand{\fwd}{forward discrepancy}
\newcommand{\bwd}{backward discrepancy}

\newcommand{\dataset}{\textsc{DepthQA}}

\NewTColorBox[auto counter]{example}{ O{!htbp} m O{} }{
  enhanced,
  breakable,
  float*,
  floatplacement={#1},
  width=\textwidth,
  colback=gray!5!white,
  colframe=gray!75!black,
  title={Example~\thetcbcounter: #2},
  enlarge top by=5mm,
  enlarge bottom by=5mm,
  label={#3}
}

\newcommand{\three}{\colorbox{green!40}{\textbf{\texttt{D3}}}}
\newcommand{\two}{\colorbox{cyan!30}{\textbf{\texttt{D2}}}}
\newcommand{\one}{\colorbox{blue!30}{{\textbf{\texttt{D1}}}}}
\newcommand{\draftonly}[1]{#1}
\renewcommand{\draftonly}[1]{}

\title{Hierarchical Deconstruction of LLM Reasoning:\\A Graph-Based Framework for Analyzing Knowledge Utilization}

\author{
    Miyoung Ko$^{1}$\thanks{Equal contribution.} \quad  Sue Hyun Park$^{1*}$ \quad  
        Joonsuk Park$^{2,3,4}$\thanks{Equal advising.} \quad Minjoon Seo$^{1}$\footnotemark[2]\\ 
        \textsuperscript{1}KAIST,
        \textsuperscript{2}NAVER AI Lab, 
        \textsuperscript{3}NAVER Cloud,
        \textsuperscript{4}University of Richmond \\
\texttt{\{miyoungko, suehyunpark, minjoon\}@kaist.ac.kr} \quad \\
\texttt{park@joonsuk.org} \quad}

\begin{document}
\maketitle

\begin{abstract}
Despite the advances in large language models (LLMs), how they use their knowledge for reasoning is not yet well understood.
In this study, we propose a method that deconstructs complex real-world questions into a graph, representing each question as a node with predecessors of background knowledge needed to solve the question. 
We develop the \dataset~dataset, deconstructing questions into three depths: (i) recalling conceptual knowledge, (ii) applying procedural knowledge, and (iii) analyzing strategic knowledge. 
Based on a hierarchical graph, we quantify \textit{\fwd}, a discrepancy in LLM performance on simpler sub-problems versus complex questions. 
We also measure \textit{\bwd} where LLMs answer complex questions but struggle with simpler ones. 
Our analysis shows that smaller models exhibit more discrepancies than larger models. 
Distinct patterns of discrepancies are observed across model capacity and possibility of training data memorization. 
Additionally, guiding models from simpler to complex questions through multi-turn interactions improves performance across model sizes, highlighting the importance of structured intermediate steps in knowledge reasoning. 
This work enhances our understanding of LLM reasoning and suggests ways to improve their problem-solving abilities.

\end{abstract}

\section{Introduction}
\label{sec:introduction}
\begin{figure}[t]
    \includegraphics[width=0.47\textwidth]{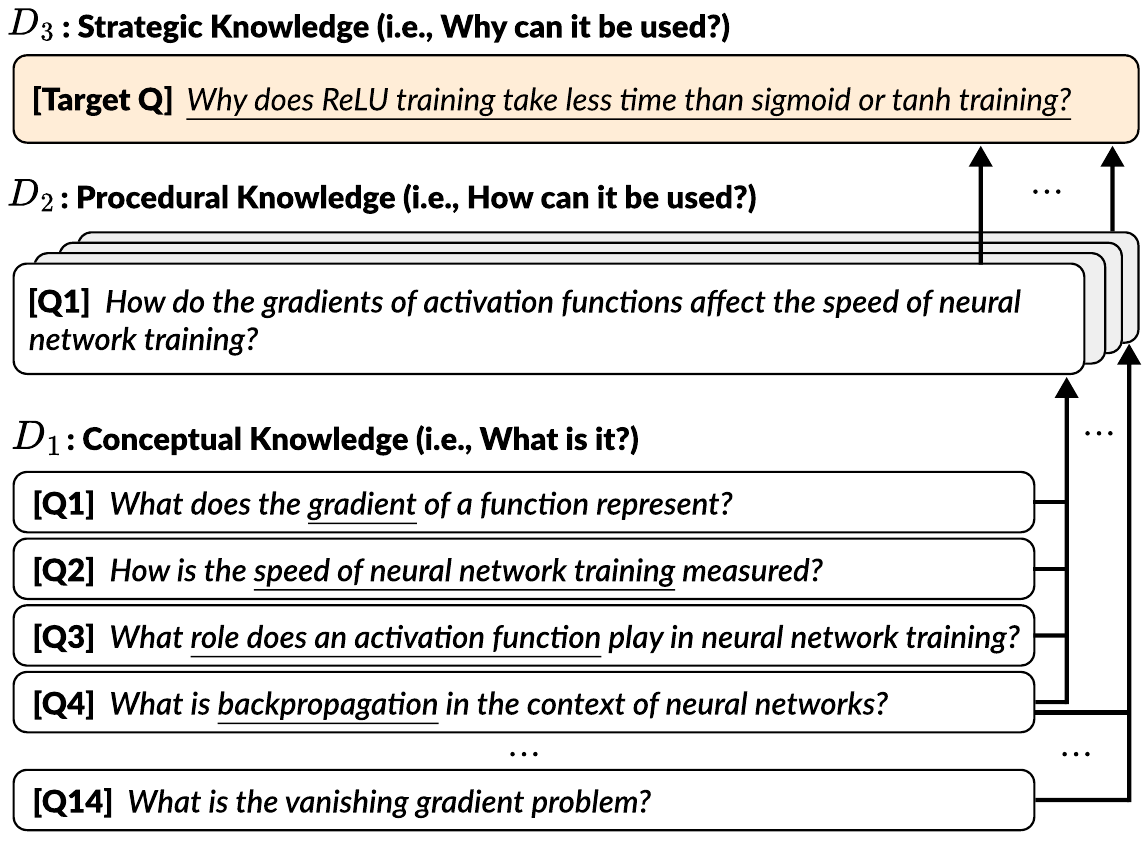}
    \vspace{-0.1cm}
    \caption{Example of reasoning across depths, showing a sequence of questions from $D_1$ (conceptual knowledge) to $D_3$ (strategic knowledge). Questions that ask deeper levels of knowledge require reasoning from multiple areas of shallower knowledge, which are represented as sub-questions.}
    \label{fig:overview}
    \vspace{-0.3cm}
\end{figure}

With the rapid advancement of Large Language Models (LLMs), research interest has increasingly centered on their reasoning capabilities, particularly in solving complex questions. 
While many studies have assessed the general reasoning capabilities of LLMs~\cite{wei2022emergent,qin-etal-2023-chatgpt, srivastava2023beyond}, the specific aspect of how these models recall and then utilize factual knowledge during reasoning has not been thoroughly explored.
Some research~\cite{NEURIPS2023_deb3c281, press-etal-2023-measuring, wang2024grokked} concentrate on straightforward reasoning tasks such as combining and comparing simple biographical facts to investigate the implicit reasoning skills of LLMs.
However, real-world questions often demand more intricate reasoning processes that cannot be easily broken down into simple factual units. 
For instance, as presented in Figure~\ref{fig:overview}, to answer ``Why does ReLU training take less time than sigmoid or tanh training?'', one must not only recall what an activation function is but also compare the characteristics of activation functions and understand the causal relationship between gradients and training speed. 
This type of reasoning requires drawing conclusions beyond simply aggregating facts.

To analyze the reasoning ability of LLMs in solving real-world questions, we propose a deconstruction of complex questions into a graph structure. 
In this structure, each node is represented by a question that signifies a specific level of knowledge. 
We adopt Webb's Depth of Knowledge~\cite{webb1997criteria, webb1999alignment, webb2002depth}, which assesses both the content and the depth of understanding required. 
Webb's Depth of Knowledge categorizes questions into three levels: mere recall of information ($D_1$), application of knowledge ($D_2$), and strategic thinking ($D_3$). 
The transition from shallower to deeper nodes involves applying the knowledge gained from shallower nodes and performing reasoning to tackle harder problems.
This approach emphasizes the gradual accumulation and integration of knowledge to address real-world problems effectively.

\begin{figure}[t]
    \includegraphics[width=0.48\textwidth]{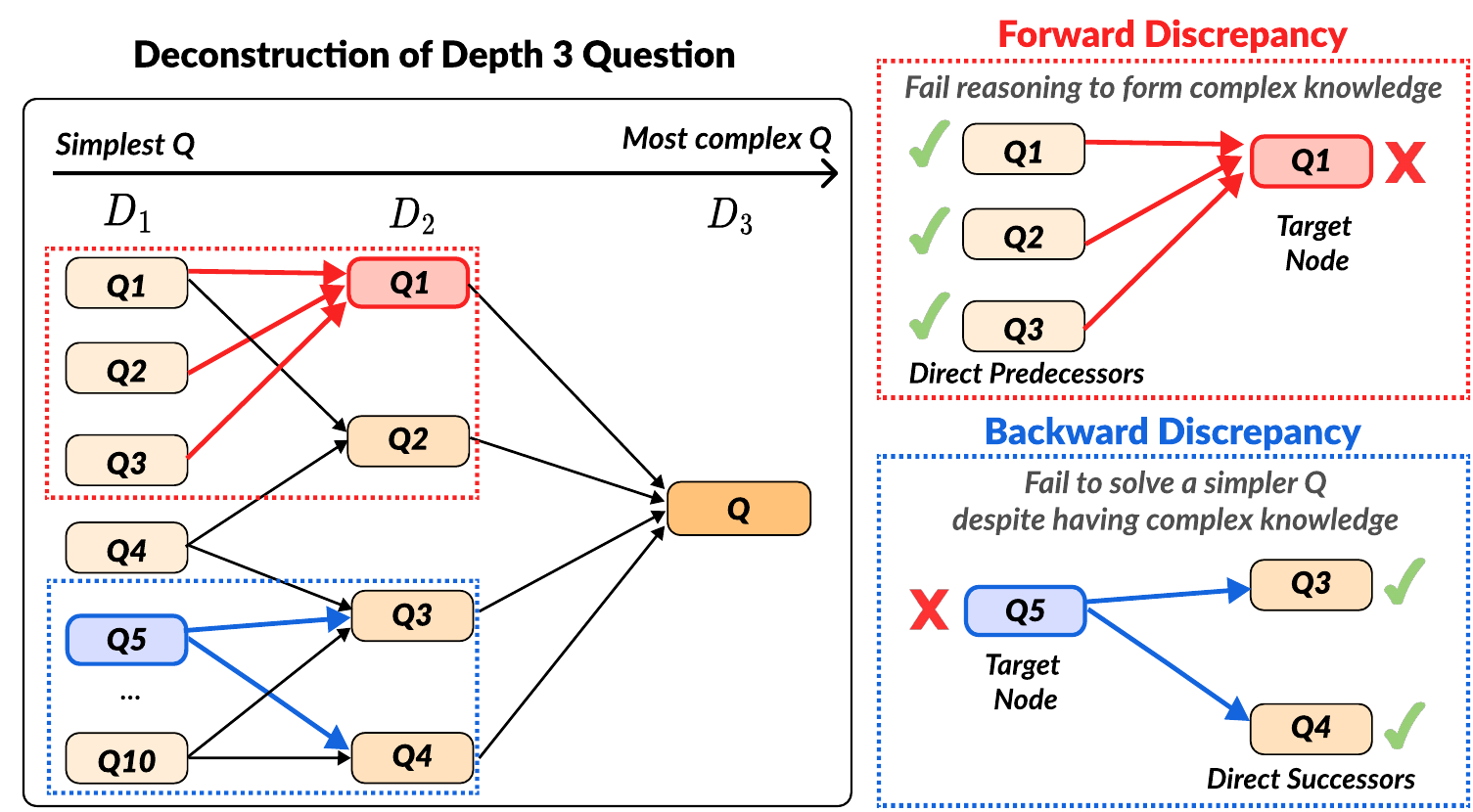}
    \vspace{-0.4cm}
    \caption{Hierarchical structure of a deconstructed $D_3$, illustrating forward and backward discrepancies. Transition to deeper nodes requires acquiring and reasoning with knowledge from the connected shallower nodes.}
    \label{fig:discrepancy}
    \vspace{-0.3cm}
\end{figure}

We introduce the resulting \dataset, a collection of deconstructed questions and answers derived from human-written, scientific $D_3$ questions in the TutorEval dataset~\cite{chevalier2024language}. 
The target complex questions are in $D_3$, and we examine the utilization of multiple layers of knowledge and reasoning in the sequence of $D_1$, $D_2$, and $D_3$. 
Figure~\ref{fig:discrepancy} illustrates how the deconstruction process results in a hierarchical graph connecting $D_1$ to $D_3$ questions. 
Based on the hierarchical structure, we first measure forward reasoning gaps, denoted as \textit{\fwd}, which are differences in LLM performance on simpler sub-problems compared to more complex questions requiring advanced reasoning. 
Additionally, we introduce \textit{\bwd}, which quantifies inconsistencies where LLMs can successfully answer complex inquiries but struggle with simpler ones. 
This dual assessment provides a comprehensive evaluation of the models' reasoning capabilities across different levels of complexity.

Using \dataset, we investigate the knowledge reasoning ability of various instruction-tuned LLMs in the LLaMA 2~\citep{touvron2023llama}, LLaMA 3~\citep{llama3modelcard}, Mistral~\citep{jiang2023mistral}, and Mixtral~\citep{jiang2024mixtral} family, varying in size from 7B to 70B. 
We compare the relationship between model capacities and depthwise discrepancies, showing that smaller models exhibit larger discrepancies in both directions.  
We further analyze how reliance on memorization of training data affects discrepancy, revealing that forward and backward discrepancies in large models originate from distinct types of failures.
Finally, to examine the importance of structured intermediate steps in reasoning, we gradually guide models from simpler to more advanced questions through multi-turn interactions, consistently improving performance across various model sizes. 

The contributions of our work are threefold:

\vspace{-.3em}
\begin{itemize}[leftmargin=1.3em]
    \setlength\itemsep{-0.1em}
     \item We propose to connect complex questions with simpler sub-questions by deconstructing questions based on depth of knowledge.
     \item We design the {\dataset} dataset to evaluate LLMs' capability to form complex knowledge through reasoning. We measure forward and backward reasoning discrepancies across different levels of question complexity.
     \footnote{We release our dataset and code at \href{https://github.com/kaistAI/knowledge-reasoning}{\texttt{github.com/kaistAI/knowledge-reasoning}}.}
     \item We investigate the reasoning abilities of LLMs with various capacities, analyzing the impact of model size and training data memorization on discrepancies. We demonstrate the benefits of structured, multi-turn interactions to perform complex reasoning.
\end{itemize}
\vspace{-.3em}

\section{Related Work}
\label{sec:related_work}
Recent advancements have highlighted the impressive reasoning abilities of transformer language models across a wide range of tasks \cite{wei2022emergent, Zhao2023ASO}. Despite the success, numerous studies have found that these models often struggle with various types of reasoning, such as commonsense and logical reasoning \cite{qin-etal-2023-chatgpt, srivastava2023beyond}. Even advanced models like GPT-4~\citep{achiam2023gpt} have been noted to struggle with implicit reasoning over their internal knowledge, especially when it comes to effectively combining multiple steps to solve compositionally complex problems \cite{talmor-etal-2020-olmpics, rogers-etal-2020-primer, allen2023physics, yang2024large, wang2024grokked}.

To address these challenges, several studies have focused on better Chain-of-Thought-style prompting or fine-tuning LLMs to verbalize the intermediate steps of knowledge and reasoning during inference \cite{nye2021show, Wei2022ChainOT, kojima2022large, wang-etal-2022-iteratively, sun2023recitationaugmented, wang2023selfconsistency, liu-etal-2023-crystal}. This approach has significantly improved performance, especially in larger models with strong generalization capabilities. 
Theoretical and empirical studies investigate the advantages of verbalizations, highlighting their role in enhancing the reasoning capabilities of language models \cite{Feng2023TowardsRT, wang-etal-2023-towards, li2024chain}. The analysis of step-by-step reasoning abilities has matured further based on ontological~\cite{saparov2023language} and mechanistic perspectives \cite{hou-etal-2023-towards, dutta2024how}.

In our proposed dataset, the most complex questions often necessitate implicit intermediate steps to reach a conclusion, which can be benefited from explicit verbalized reasoning. 
However, unlike previous works, our setup does not induce detailed step-by-step explanation \textit{contained} in an answer to a complex question. Instead, we represent intermediate steps for a complex question in the form of \textit{sub-questions} and gather answers to every sub-question, testing a model's understanding of intermediate knowledge individually.
Our approach is similar to strategic question answering with intermediate answers~\citep{geva-etal-2021-aristotle, press-etal-2023-measuring}, but we further ensure a hierarchy of decompositions based on knowledge complexity. This allows examining discrepancies between questions of varying complexities, providing a distinct assessment of multi-step reasoning abilities.

Another line of work focuses on understanding transformers' knowledge and reasoning through controlled experiments \cite{NEURIPS2022_77c6ccac, akyurek2023what, dai-etal-2023-gpt, Oswald2022TransformersLI, prystawski2023why, feng2024how}. Numerous studies on implicit reasoning often aim to identify latent reasoning pathways, but most have focused on simple synthetic tasks or toy models \cite{nanda2023progress, NEURIPS2023_34e1dbe9, hou2023towards}, or evaluating through binary accuracy of short-form model predictions without considering intermediate steps \cite{yang2024large, wang2024grokked}.
Our \dataset, in contrast, challenges a model to answer complex \textit{real-world} questions that require \textit{diverse} reasoning types in \textit{long-form} text. 
\dataset~further requires diverse types of reasoning across different depths, such as inductive and procedural reasoning, in addition to the comparative and compositional reasoning explored in prior studies \citep{press-etal-2023-measuring, allen2023physics, wang2024grokked}. 
This approach provides a more practical and nuanced assessment of the model's reasoning capabilities.

\section{Graph-based Reasoning Framework}
\label{sec:dataset}
We develop a novel graph-based representation that delineates the dependencies between different levels of knowledge. We represent nodes as questions (Section~\ref{subsec:depth-definition}) and edges as reasoning processes (Section~\ref{subsec:decomposition-criteria}). Based on the graph definition, we construct a dataset that encompasses diverse concepts and reasoning types (Section~\ref{subsec:data-construction}).

\subsection{Knowledge Depth in Nodes}
\label{subsec:depth-definition}

We represent each node as a question tied to a specific layer of knowledge. As our approach to addressing real-world problems emphasizes the \textit{gradual} accumulation of knowledge similar to educational goals, we adopt the Webb's Depth of Knowledge (DOK)~\citep{webb1997criteria, webb1999alignment, webb2002depth} widely used in education settings to categorize the level of questions. The depth of knowledge levels $D_k (k \in \{1,2,3\})$\footnote{We exclude the highest level in the original Webb's DOK, $D_4$, as this level often includes interactive or creative activities and is rare or even absent in most standardized assessment~\citep{webb2002depth, hess2006applying}.} in questions are defined as follows:

\begin{enumerate}
    \renewcommand{\labelenumi}{$D_\arabic{enumi}$.}
    \item \textbf{Factual and conceptual knowledge}: The question involves the acquisition and recall of information, or following a simple formula, focusing on \textit{what} the knowledge entails.
    \item \textbf{Procedural knowledge}: The question necessitates the application of concepts through the selection of appropriate procedures and step-by-step engagement, concentrating on \textit{how} the knowledge can be utilized.
    \item \textbf{Strategic knowledge}: The question demands analysis, decision-making, or justification to address non-routine problems, emphasizing \textit{why} the knowledge is applicable.
\end{enumerate}

The levels can be viewed as \textit{ceilings} that establish the extent or depth of an assessee's understanding~\citep{hess2006applying}, a concept recognized as a valuable assessment tool in educational contexts~\citep{hess2009cognitive}. Accordingly, we associate simpler questions with shallower depths and more complex questions with deeper depths.

\subsection{Criteria for Reasoning in Edges}
\label{subsec:decomposition-criteria}
To conceptualize how simpler knowledge contributes to the development of complex knowledge, we define edges in our framework as transitions from a node at $D_k$ to at least one direct successor node at $D_{k+1}$\footnote{A foundational concept may apply to multiple advanced questions.}. We perceive that advancing to deeper knowledge often requires synthesizing multiple aspects of simpler knowledge. Thus, a $D_{k}$ node should connect to multiple direct predecessor $D_{k-1}$ nodes. This configuration establishes hierarchical dependencies among $D_1$, $D_2$, and $D_3$ questions, effectively modeling the progression needed to deepen understanding and engage with higher-order knowledge (See graph in Figure~\ref{fig:discrepancy}). Additionally, we establish three criteria to ensure that edges accurately represent the reasoning processes from shallower questions.

\begin{enumerate}
    \renewcommand{\labelenumi}{\textbf{C\arabic{enumi}}.}
    \item \textbf{Comprehensiveness:} Questions at lower levels should aim to cover all foundational concepts necessary to answer a question at higher levels. This ensures that no critical knowledge gaps exist as the complexity increases.
    \item \textbf{Implicitness:} Questions at lower levels should avoid directly revealing answers or heavily hinting at solutions for higher-level questions. This encourages independent reasoning relying on the synthesis of implicit connections between nodes rather than straightforward clues. 
    \item \textbf{Non-binary questioning:} Questions should elicit detailed, exploratory responses instead of simple yes/no answers. Given that LLMs may have an inherent positivity bias which leads them to prefer affirmative responses~\citep{augustine2011positivity, dodds2015human, lesswrong2023biased}, this helps in evaluating deep reasoning abilities beyond superficial or biased reasoning.
\end{enumerate}

\subsection{Dataset: {\dataset}}\label{subsec:data-construction}
We create {\dataset}, a new question answering dataset for testing graph-based reasoning. The dataset is constructed in a top-down approach, deconstructing $D_3$ nodes into $D_2$ nodes, then into $D_1$, creating multiple edges at each step (Table~\ref{table:data-stats}). We design the construction process to meticulously backtrack the knowledge necessary for complex questions while meeting our three criteria for reasoning transition representation. 

\paragraph{$D_3$ question curation}
We select real-world questions from the TutorEval~\citep{chevalier2024language} dataset, which contains human-crafted queries based on college-level mathematical and scientific content from textbooks\footnote{Textbooks are designed with a scaffolding approach to knowledge development.} available on \href{https://libretexts.org/}{\texttt{libretexts.org}}. 
Note that while these textbooks may be part of models' pre-training data due to online availability, TutorEval's human-written questions challenge models to generalize familiar concepts beyond direct training examples.
We procure only complex $D_3$ questions from TutorEval, sorting them out using GPT-4 Turbo\footnote{We use the \href{https://platform.openai.com/docs/models/gpt-4-turbo-and-gpt-4}{\texttt{gpt-4-0125-preview}} version for GPT-4 Turbo throughout this work, including data construction, verification, and experiments.}~\citep{achiam2023gpt} with guidance on depth of knowledge levels. 
From an initial set of 834 questions, we manually refine our selection to 91 self-contained $D_3$ questions, ensuring clarity. 
We use GPT-4 Turbo to generate reference answers for each TutorEval question\footnote{\citet{chevalier2024language} reports that GPT-4 excels in solving TutorEval problems with 92\% correctness.}, based on the original context and the model's self-annotated depth of knowledge. These reference answers are guided by the ground-truth key points provided by the author of each question.

\begin{table}[t]
\centering
\small
\resizebox{0.49\textwidth}{!}{
\begin{tabular}{lrrrrrrr}
    \toprule
    \multirow{2}{*}{Domain} &\multicolumn{3}{c}{\# Questions} &\multicolumn{4}{c}{\# Edges between questions} \\
    \cmidrule(lr){2-4} \cmidrule(lr){5-8} 
    &$D_1$ &$D_2$ &$D_3$ &$D_1 \rightarrow D_2$ &$D_2 \rightarrow D_3$  \\
    \toprule
    Math &573 &193 &49 &774 &196 \\
    Computer Science &163 &54 &14 &212 &55 \\
    Environmental Science &147 &44 &11 &175 &44 \\
    Physics &140 &40 &10 &154 &40 \\
    Life Sciences &98 &28 &7 &111 &28 \\
    Math $\rightarrow$ \{CS, Physics\}  &- &- &- &11 &0 \\
    \midrule
    Total &1,121 &359 &91 &1,437 &363 \\
    \bottomrule
\end{tabular}
}
\caption{Statistics of {\dataset}.}
\label{table:data-stats}
\vspace{-0.3cm}
\end{table}

\begin{table*}[t]\centering
\small
\resizebox{0.96\textwidth}{!}{
\begin{tabular}{clp{0.78\textwidth}r}\toprule
Depth &Reasoning type &Example question &\% \\\midrule
\multirow{7}{*}{3} &Comparative &\textit{In the context of computer programming, what is the difference between for and while, are they always exchangeable? Show me some cases where one is strongly preferred to the other.} &21.1 \\
\cmidrule{2-4}
&Causal &\textit{How does deflection of hair cells along the basilar membrane result in different perceived sound frequences?} &10.5 \\
\cmidrule{2-4}
&Inductive &\textit{How could a process satisfying the first law of thermodynamics still be impossible?} &8.8 \\
\cmidrule{2-4}
&Criteria Development &\textit{Explain if a matrix always have a basis of eigenvectors.} &8.8 \\
\midrule
\multirow{5}{*}{2} &Relational &\textit{What factors influence the time complexity of searching for an element in a data structure?} &22.6 \\
\cmidrule{2-4}
&Procedural &\textit{Describe the process involved in solving cubic equations using the cubic formula.} &13.4 \\
\cmidrule{2-4}
&Application &\textit{How can sustainable agricultural practices contribute to food security and economic development in developing countries?} &7.3 \\
\bottomrule
\end{tabular}
}
\caption{Representative examples of required reasoning skills in $D_3$ and $D_2$. \% of instances within each depth that include the reasoning type is reported. Note that multiple reasoning types can be included in a single question.}\label{tab:reasoning_type}
\vspace{-0.3cm}
\end{table*}

\paragraph{Question deconstruction}
For each $D_k$ question, we use GPT-4 Turbo to generate up to four $D_{k-1}$ questions. The prompt includes definitions for all three knowledge depths and decomposition examples to guide the deconstruction process. We provide $D_k$ with its reference answer to ensure extracted knowledge remains relevant for more challenging questions, adhering to \textbf{C1 (Comprehensiveness)}.  
We decide the optimal number of decompositions to four based on qualitative analysis, balancing comprehensiveness and implicitness: outlining every implicit reasoning step enhances comprehensiveness but may reduce implicitness, and vice versa.  Our prompt instructions carefully address this tradeoff to satisfy \textbf{C2 (Implicitness)}.

\paragraph{Deduplication and question augmentation}
We identified redundancies in knowledge and reasoning processes, where similar content appeared across different $D_1$ nodes linked to the same $D_2$ node, or between unconnected $D_1$ and $D_2$ nodes (example in Table~\ref{tab:appendix_dedupe}). To address this, we utilize a Sentence Transformers embedding model\footnote{\href{https://huggingface.co/sentence-transformers/all-mpnet-base-v2}{\texttt{sentence-transformers/all-mpnet-base-v2}}}~\citep{reimers-gurevych-2019-sentence} to detect and remove near-duplicate questions based on cosine similarity of their embeddings. We then employ GPT-4 Turbo to generate new, targeted questions and answers, filling any gaps in knowledge coverage. This approach has reduced misclassification of $D_1$ questions as $D_2$ by 88\%, markedly enhancing \textbf{C2 (Implicitness)}. It has also decreased the total number of near-duplicates by decreased by 88\%, further improving \textbf{C1 (Comprehensiveness)}. We subsequently update our graph data structure with these modifications.

\paragraph{Question debiasing}
Lastly, we undertake the task of manually rewriting 53 questions that originally invoke binary ``yes'' or ``no'' answers, ensuring \textbf{C3 (Non-binary Questioning)}. For example, a question that begins with ``If I understand correctly...'' is transformed into ``Clarify my understanding that...'', prompting the model to directly engage in analytical thinking rather than relying on simple affirmations or negations of the correctness.

\paragraph{Verification of hierarchy}
We conduct human annotation to verify the three criteria that shapes the reasoning hierarchy, reporting positive results in Appendix~\ref{appendix:data_quality_check}. On 27.5\% of {\dataset}, an average of 83.5\% of relations are fully comprehensive and 89.5\% of sub-questions are fully implicit, with 98.7\% of all questions being non-binary. Further details and examples in the construction process are in Appendix~\ref{appendix:dataset}. Prompts are in Appendix~\ref{appendix:data_prompts}.

\subsection{Diversity of Reasoning Processes}\label{subsec:data-analysis}

Using a sample of 20 $D_3$ questions along with their interconnected 80 $D_2$ and 320 $D_1$ questions, we examine the types of reasoning needed to progress from basic to complex knowledge levels. We discover that nearly all questions necessitate the identification and extraction of several pieces of relevant information to synthesize comprehensive answers. Table~\ref{tab:reasoning_type} displays examples of questions requiring advanced reasoning skills, such as interpreting relationships between concepts, applying specific conditions, and handling assumptions, demonstrating that basic knowledge manipulation is insufficient. This diversity in reasoning types within our dataset robustly challenges LLMs to demonstrate sophisticated cognitive abilities. Detailed statistics and additional examples of reasoning types are provided in Appendix~\ref{appendix:analysis}.


\section{Experiments}
\label{sec:experiments}
In this section, we present experiments on the depthwise reasoning ability of LLMs using {\dataset}. 
We first explain the evaluation metrics and models (Section~\ref{subsec:setup}). 
Experimental results that follow are overall depthwise and discrepancy evaluation results (Section~\ref{subsec:depthwise_results}), the impact of memorization in knowledge reasoning (Section~\ref{subsec:memorization}), and the effect of enforcing knowledge-enhanced reasoning via multi-turn inputs or prompt inputs (Section~\ref{subsec:effect_enforcing}).

\subsection{Experiment Setup}\label{subsec:setup}

\paragraph{Depthwise evaluation}
For each question $q_k$ with depth $k$ ($D_k$), we score the factual correctness of the predicted answer on a scale from 1 to 5. We employ the LLM-as-a-Judge approach, which correlates highly with human judgments in scoring long-form responses~\citep{zheng2024judging, kim2024prometheus, lee2024prometheus, kim2024prometheus2}. Specifically, we utilize GPT-4 Turbo \citep{achiam2023gpt} for absolute scoring. Following \citet{kim2024prometheus} and~\citet{lee2024prometheus}, the model generates a score and detailed feedback for each question, reference answer, and prediction based on a defined scoring rubric. Further details on the evaluation process are provided in Appendix~\ref{appendix:experiments}. The exact input prompt for the LLM judge including the accuracy score rubric is in Appendix~\ref{appendix:evaluation_prompts}. The reliability of the LLM evaluation results in our setting is evidenced by high annotation agreement with human evaluations, as explained in Appendix~\ref{appendix:reliability_llm_judge}. 
We report \textbf{average accuracy} at $D_k$, the averaged factual correctness of questions at depth $k$.

\begin{table*}[t]\centering
\resizebox{\textwidth}{!}{
\begin{tabular}{lcccccccccc}\toprule
\multirow{2}{*}{Model} &\multicolumn{4}{c}{\textbf{Average Accuracy} $\uparrow$} &\multicolumn{3}{c}{\textbf{Forward Discrepancy} $\downarrow$} &\multicolumn{3}{c}{\textbf{Backward Discrepancy} $\downarrow$} \\\cmidrule(lr){2-5}\cmidrule(lr){6-8}\cmidrule(lr){9-11}
&\textbf{$D_1$} &\textbf{$D_2$} &\textbf{$D_3$} & Overall &$D_2 \rightarrow D_3$ &$D_1 \rightarrow D_2$ & Overall &$D_2 \rightarrow D_3$ &$D_1 \rightarrow D_2$ & Overall \\\toprule
LLaMA 2 7B Chat &3.828 &3.320 &3.165 &3.673 &\cellcolor[HTML]{f0b4b4}0.130 &\cellcolor[HTML]{e78787}0.181 &\cellcolor[HTML]{e88c8c}0.176 &\cellcolor[HTML]{6e9fec}0.219 &\cellcolor[HTML]{c7daf8}0.110 &\cellcolor[HTML]{b4cdf5}0.134 \\
LLaMA 2 13B Chat &4.289 &3.872 &3.615 &4.155 &\cellcolor[HTML]{eca0a0}0.152 &\cellcolor[HTML]{eb9b9b}0.158 &\cellcolor[HTML]{eb9c9c}0.157 &\cellcolor[HTML]{bbd2f6}0.126 &\cellcolor[HTML]{e2ecfb}0.078 &\cellcolor[HTML]{dae7fa}0.088 \\
LLaMA 2 70B Chat &4.495 &4.153 &4.022 &4.390 &\cellcolor[HTML]{f1b7b7}0.126 &\cellcolor[HTML]{efafaf}0.136 &\cellcolor[HTML]{efb0b0}0.134 &\cellcolor[HTML]{b2ccf5}0.136 &\cellcolor[HTML]{eff4fd}0.063 &\cellcolor[HTML]{e2ecfb}0.079 \\
Mistral 7B Instruct v0.2 &4.280 &3.897 &4.000 &4.176 &\cellcolor[HTML]{f7d5d5}0.092 &\cellcolor[HTML]{eb9d9d}0.157 &\cellcolor[HTML]{eda5a5}0.147 &\cellcolor[HTML]{acc8f4}0.144 &\cellcolor[HTML]{e9f0fc}0.070 &\cellcolor[HTML]{dae7fa}0.088 \\
Mixtral 8x7B Instruct v0.1 &\ul{4.599} &\ul{4.532} &\ul{4.429} &\ul{4.574} &\cellcolor[HTML]{f8d9d9}0.087 &\cellcolor[HTML]{f9e0e0}0.079 &\cellcolor[HTML]{f9dfdf}0.081 &\cellcolor[HTML]{eff5fd}0.063 &\cellcolor[HTML]{eef4fd}\ul{0.063} &\cellcolor[HTML]{eef4fd}0.063 \\
LLaMA 3 8B Instruct &4.482 &4.351 &4.286 &4.440 &\cellcolor[HTML]{f8dddd}\ul{0.083} &\cellcolor[HTML]{f6d2d2}0.096 &\cellcolor[HTML]{f7d4d4}0.093 &\cellcolor[HTML]{dae7fa}0.088 &\cellcolor[HTML]{e8f0fc}0.072 &\cellcolor[HTML]{e5eefc}0.075 \\
LLaMA 3 70B Instruct &\textbf{4.764} &\textbf{4.749} &\textbf{4.648} &\textbf{4.754} &\cellcolor[HTML]{fcecec}\textbf{0.065} &\cellcolor[HTML]{fef9f9}\textbf{0.050} &\cellcolor[HTML]{fef7f7}\textbf{0.053} &\textbf{0.043} &\cellcolor[HTML]{feffff}\textbf{0.044} &\textbf{0.044} \\
GPT-3.5 Turbo &4.269 &4.251 &4.011 &4.250 &\cellcolor[HTML]{f5cece}0.100 &\cellcolor[HTML]{fae6e6}\ul{0.072} &\cellcolor[HTML]{f9e1e1}\ul{0.078} &\cellcolor[HTML]{fdfeff}\ul{0.046} &\cellcolor[HTML]{ebf2fd}0.067 &\cellcolor[HTML]{eff5fd}\ul{0.063} \\
\bottomrule
\end{tabular}}
\caption{Depthwise reasoning performance of large language models. \textbf{Bold} indicates the best-performing model, and \ul{underline} represents the second best performance. A darker color indicates a higher discrepancy.}
\label{table:llm_performance}
\vspace{-0.3cm}
\end{table*}

\paragraph{Discrepancy evaluation}
As we deconstruct complex questions into a hierarchical graph, we can measure \textit{\fwd}~and \textit{\bwd}~between neighboring questions. \textbf{\Fwd}~measures the differences in performance on sub-problems compared to deeper questions requiring advanced reasoning. Given a question $q_k$  at $D_k \in \{2, 3\}$,  let $DP(q_k)$  represents a set of direct predecessor questions at $D_{k-1}$. Then \fwd~for $q_k$ is defined as follows:
\begin{multline}
\text{Forward Discrepancy}(q_k) = \\ \max \left( 0, \frac{1}{4} \left( \text{avg}_{q \in DP(q_k)} [f(q)] - f(q_k) \right) \right)
\end{multline}
\noindent where $f$ is a function of a question that outputs factual correctness, as measured by an LLM evaluator. \textbf{\Bwd}, conversely, quantifies inconsistencies where LLMs can successfully answer deeper questions but struggle with shallower ones. Given a question $q_k$  with $D_k \in \{1, 2\}$, let $DS(q_k)$  represent a set of direct successor questions at $D_{k-1}$. Then \bwd~is defined as follows:
\begin{multline}
\text{Backward Discrepancy}(q_k) = \\
\max \left( 0, \frac{1}{4} \left( \text{avg}_{q \in DS(q_k)} [f(q)] - f(q_k) \right) \right).
\end{multline}
Both \fwd~and \bwd~are normalized to the range [0, 1] by dividing by the \textit{maximum possible score gap}, which is 4 at our scoring range from 1 to 5. 
To highlight gaps across depths, we set a strict accuracy threshold of 4 and report the average discrepancies only for examples where the mean score for $DP(q^k)$  and  $DS(q^k)$ exceeds this threshold. This excludes cases where models perform poorly at both depths.

\paragraph{Models}\label{paragraph:models}
We mainly probe into the depthwise knowledge reasoning ability of open-source LLMs.
We test representative open-source models based on the LLaMA~\citep{touvron2023llama} architecture, including LLaMA 2 \{7B, 13B, 70B\} Chat~\citep{touvron2023llama}, Mistral 7B Instruct v0.2~\citep{jiang2023mistral}, Mixtral 8x7B Instruct v0.1~\citep{jiang2024mixtral}, and LLaMA 3 \{8B, 70B\} Instruct~\citep{llama3modelcard}. 
Additionally, we include the latest GPT-3.5 Turbo\footnote{\href{https://platform.openai.com/docs/models/gpt-3-5-turbo}{\texttt{gpt-3.5-turbo-0125}}}~\citep{chatgpt} to compare the performance of these open-source models against a proprietary model.

\subsection{Depthwise Knowledge Reasoning Results}\label{subsec:depthwise_results}
\paragraph{Larger models exhibit smaller discrepancies.}
Table~\ref{table:llm_performance} presents the overall depthwise reasoning performance of LLMs. As anticipated, solving questions at $D_3$ is the most challenging, showing the lowest average accuracy for all models. 
LLaMA 3 70B Instruct demonstrates the best performance across all depths, with Mixtral 8x7B Instruct achieving the second-best results.
LLaMA 3 70B Instruct also exhibits the lowest discrepancies for both forward and backward discrepancy metrics, effectively answering questions at all depths with minimal discrepancies. 
Conversely, the least capable model, LLaMA 2 7B Chat, shows the lowest average accuracy along with the highest forward and backward discrepancies. 
Note that the relatively low \fwd~from $D_1 \rightarrow D_2$ for LLaMA 2 7B Chat is due to its low performance at $D_2$.
This observation highlights the varying capabilities of different LLMs in handling questions at different depths and the inconsistencies in reasoning across depths. 

\begin{figure*}[ht!]
    \centering
    \subfloat[LLaMA 2 7B Chat]{\includegraphics[width=0.25\textwidth]{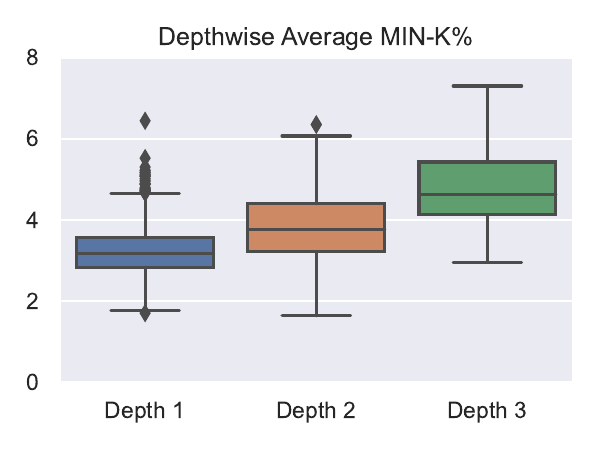}}
    \subfloat[LLaMA 2 70B Chat]{\includegraphics[width=0.25\textwidth]{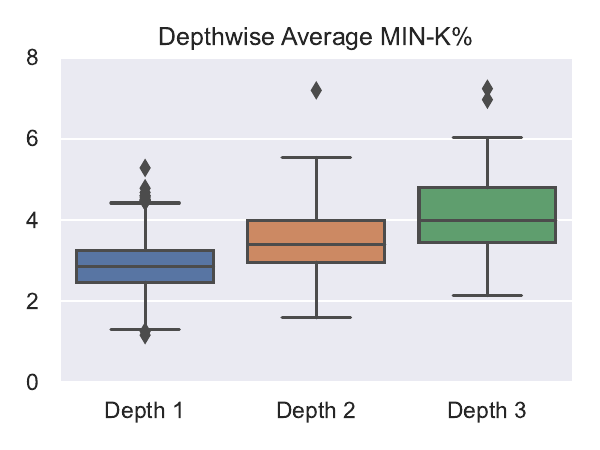}}
    \subfloat[LLaMA 3 8B Instruct]{\includegraphics[width=0.25\textwidth]{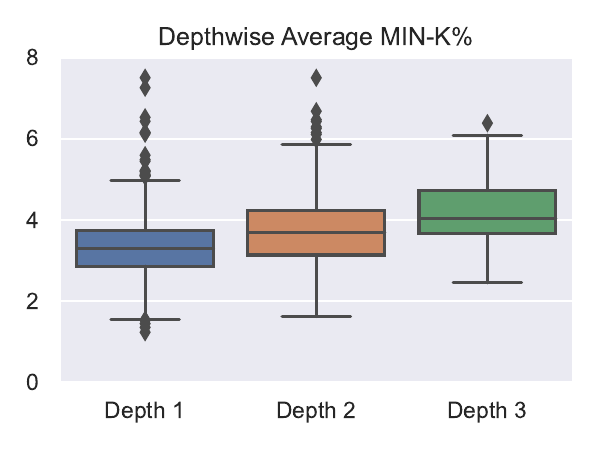}}
    \subfloat[LLaMA 3 70B Instruct]{\includegraphics[width=0.25\textwidth]{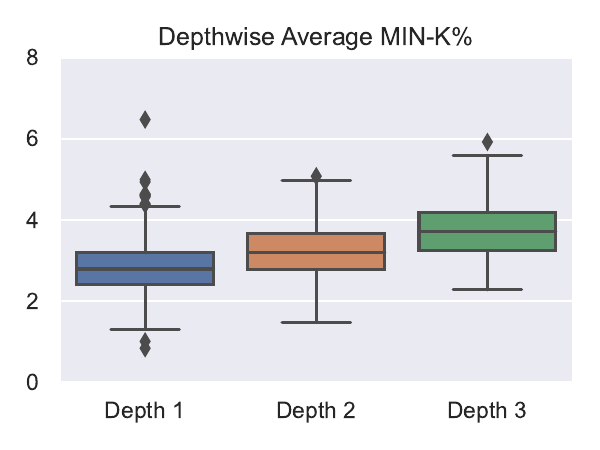}}
    \vspace{-0.1em}
    \subfloat[LLaMA 2 7B Chat]{\includegraphics[width=0.3\textwidth]{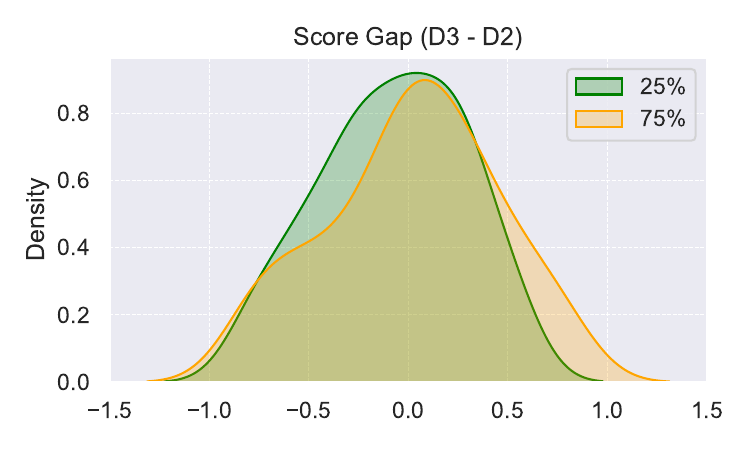}}
    \subfloat[LLaMA 2 70B Chat]{\includegraphics[width=0.3\textwidth]{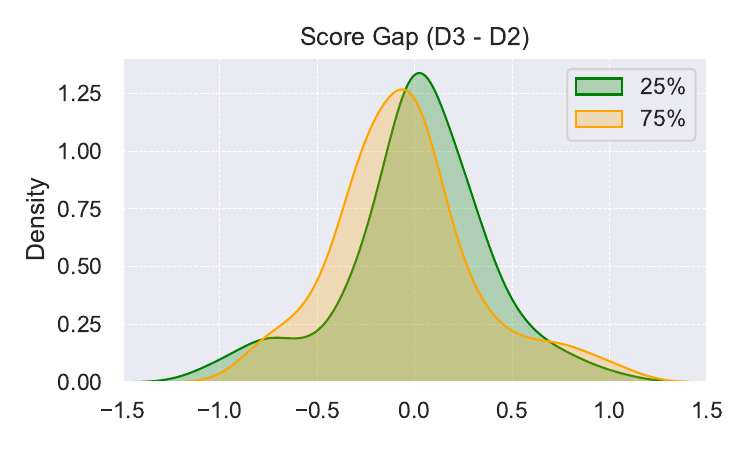}}
    \subfloat[LLaMA 3 70B Instruct]{\includegraphics[width=0.3\textwidth]{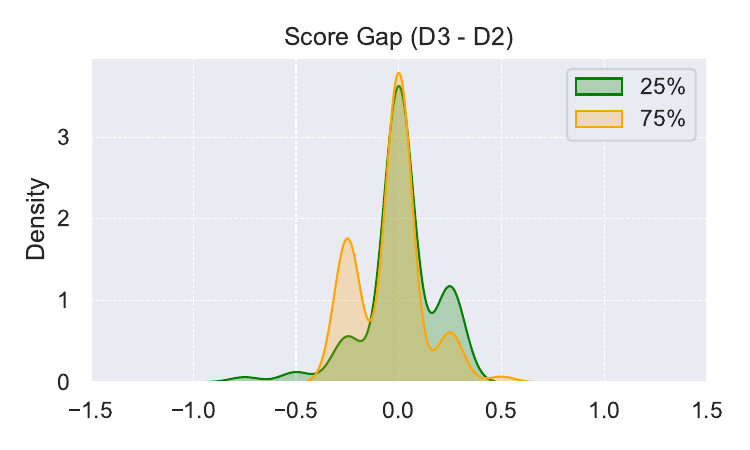}}
    \caption{Memorization analysis with Min-K\% probability. 
    (a)-(d) show the distribution of average Min-K\% probabilities at each depth. 
    (e)-(g) present the distribution of score differences between neighboring questions, whose Min-K\% probability is in the bottom 25\% or top 75\%. 
    A positive gap indicates backward discrepancy, while a negative gap represents forward discrepancy.}
    \label{fig:memorization}
    \vspace{-0.2cm}
\end{figure*}
\paragraph{Contrasting patterns of discrepancies}
We observe distinct patterns when analyzing forward and backward discrepancies separately. 
These discrepancies can be understood as a product of intensity (the magnitude of the discrepancies) and frequency (the proportion of questions showing a positive discrepancy). 
Frequency indicates how often \fwd~or \bwd~occurs, while intensity reflects the strength of the discrepancy when it happens.
Our analysis shows that \fwd~tends to occur more frequently but with lower intensity. 
For example, LLaMA 3 8B Instruct exhibits an intensity of 0.225 with a frequency of 41.44\%. 
In contrast, \bwd~is less common but has a higher intensity when they appear. 
Specifically, LLaMA 3 8B Instruct shows an intensity of 0.323 with a frequency of 23.32\% for backward discrepancies. The intensity and frequency for all models are provided in Appendix~\ref{appendix:discrepancy}.

\subsection{Memorization in Depthwise Knowledge Reasoning}\label{subsec:memorization}

\subsubsection{Depthwise Memorization}
To determine whether solving complex questions requires reasoning rather than memorization of training data, we use a pre-training data detection method to approximate potential aspects of memorization.
Following \citet{Shi2023DetectingPD}, we compare the \textbf{Min-K\% probability} within models. 
Higher values suggest a smaller possibility of predictions directly existing in the training data.
To elaborate, Min-K\% probability is calculated by averaging the negative log-likelihood of the K\% least probable tokens in the model's predictions. 
In the case where a given prediction was seen during training, outlier words with low probabilities would appear less frequently, resulting in high probabilities for the K\% tokens. 
Since Min-K\% probability is the average negative log-likelihood of such tokens, the resulting value would be lower in this case.
\footnote{For our calculations, we set $k$ to 20 and use a sequence length of 128.}

\paragraph{Models rely less on memorization for complex questions.}

Figure~\ref{fig:memorization} (a)-(d) presents the depthwise average of the Min-K\% probability for four models. 
We observe that as the depth increases, the Min-K\% probability also increases for all models. 
This indicates that answering questions based on simple conceptual knowledge corresponding to $D_1$ is more likely to be solved by recalling training data.
While shallow questions ($D_1$) can be addressed through memorization, solving deeper questions ($D_3$) requires more than just recalling a single piece of memorized knowledge, indicating a need for genuine reasoning capabilities.

\subsubsection{Memorization Gap between Depths}\label{subsubsec:memorization_gap}
Further analysis of questions in the bottom 25\% and top 75\% quantiles of the Min-K\% probability distribution provides additional insights. Note that questions in the top 75\% quantile are more likely to appear in the training data, while those in the bottom 25\% are less likely.
Figure~\ref{fig:memorization} (e)-(g) shows the score difference between neighboring questions ($D_2 \rightarrow D_3$) whose Min-K\% probability is in the bottom 25\% or top 75\%. 
We calculate the \textbf{memorization gap} as the difference between the factual correctness of $D_3$ and $D_2$, normalized by the maximum gap of 4. 
A positive value indicates higher factual accuracy for the deeper questions, signifying backward discrepancy, while a negative value indicates higher accuracy for the shallower question, representing forward discrepancy.

\paragraph{Variance of gaps}
We observe that the model with the smallest capacity, LLaMA 2 7B Chat, exhibits large variances in both positive and negative directions, showing significant forward and backward discrepancies. 
In contrast, models with larger capacities, such as LLaMA 2 70B Chat and LLaMA 3 70B Instruct, demonstrate smaller variances. 

\paragraph{Potential causes of discrepancies}
Additionally, models with larger capacities tend to show relatively higher forward discrepancies---distribution concentrated on the negative side---for the top 75\% examples, which rely less on memorization. 
On the other hand, the bottom 25\% distribution is concentrated on positive values, indicating relatively more backward discrepancies. 
This suggests that as model capacity increases, failures in knowledge reasoning result in forward discrepancies, while failures due to reliance on memorization may lead to backward discrepancies. 
The depthwise Min-K\% probability and score difference for other models are provided in Appendix~\ref{appendix:min_k_prob}.

\subsection{Qualitative Analysis of Backward Discrepancy}
To better understand how the more abnormal inconsistency---{\bwd}---can emerge, we qualitatively analyze backward discrepancy cases from the weakest model in our experiments, LLaMA 2 7B chat, and the strongest model in our experiments, LLaMA 3 70B Instruct. The examples we refer to in the following paragraphs are listed in Appendix~\ref{appendix:discrepancy_examples}.

We observe that backward discrepancies often stem from the models' ability to articulate high-level concepts but struggle with translating this understanding into precise, step-by-step procedures, particularly when mathematical operations are involved. 
This is illustrated in Example~\ref{example:appendix_common_bwd}, where both models explain the importance of continued fraction representation for tangle numbers well ($D_3$) but fail to accurately describe the process of constructing a tangle for a given number ($D_2$).

In {\bwd} cases, answers to deeper questions are more likely to be text-based and conceptual, making them easier for models to memorize that data. In contrast, shallower questions require execution of mathematical or logical operations, where the variability in the elements makes answers harder to memorize verbatim. This elucidates memorization effects on {\bwd} analyzed in Section~\ref{subsubsec:memorization_gap}.

Interestingly, we also observe how the degree of memorization contributing to backward discrepancy can vary with model capacity. Example~\ref{example:appendix_small_bwd} shows LLaMA 2 7B Chat accurately reasoning about time complexity ($D_3$) but introducing non-standard terminology for specific operations ($D_2$), suggesting the model's struggle with precise recall of basic concepts. Conversely, Example~\ref{example:appendix_large_bwd} demonstrates LLaMA 3 70B Instruct correctly recalling a complex formula ($D_3$) but failing to apply it practically ($D_2$). This indicates that the model can extensively memorize information but still struggle with its flexible application. This observation exemplifies why variance of memorization gaps can differ by model capacity, as described in Section~\ref{subsubsec:memorization_gap}.

\begin{figure} [t]
    \centering
    \subfloat[Depth 1]
    {\includegraphics[width=0.43\textwidth]
    {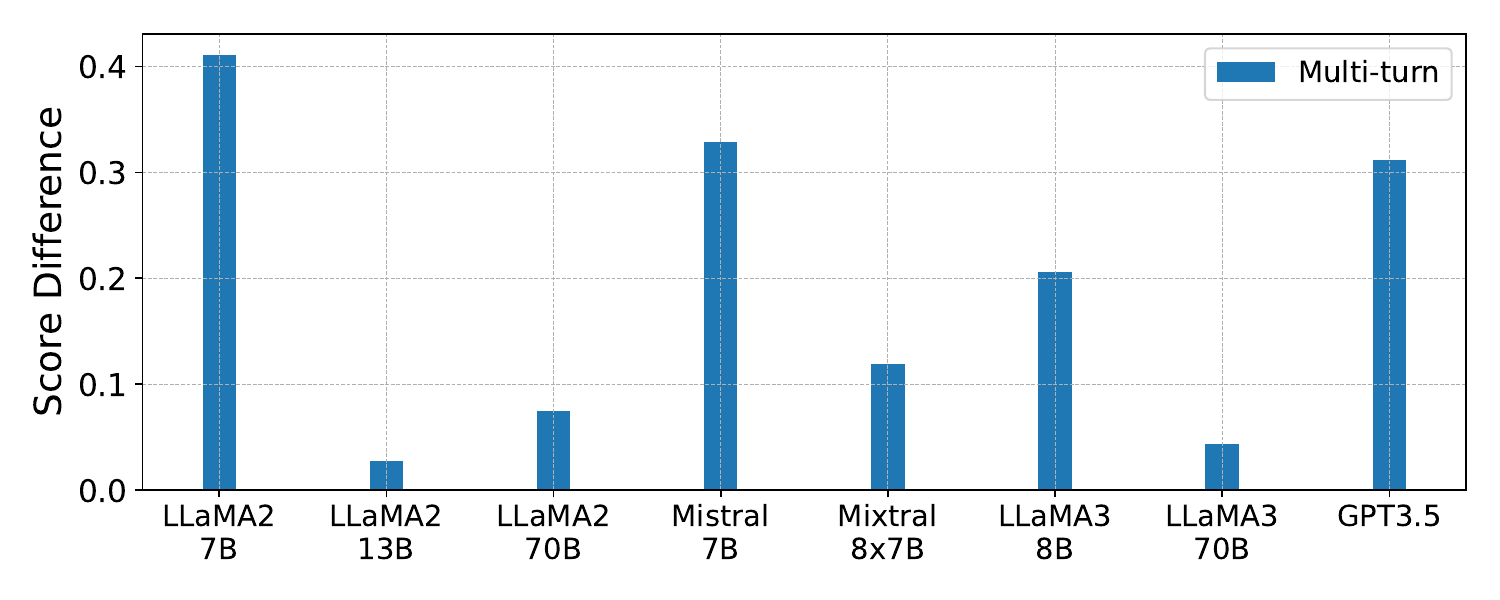}} \\
    \vspace{-0.2cm}
     \subfloat[Depth 2]     {\includegraphics[width=0.43\textwidth]{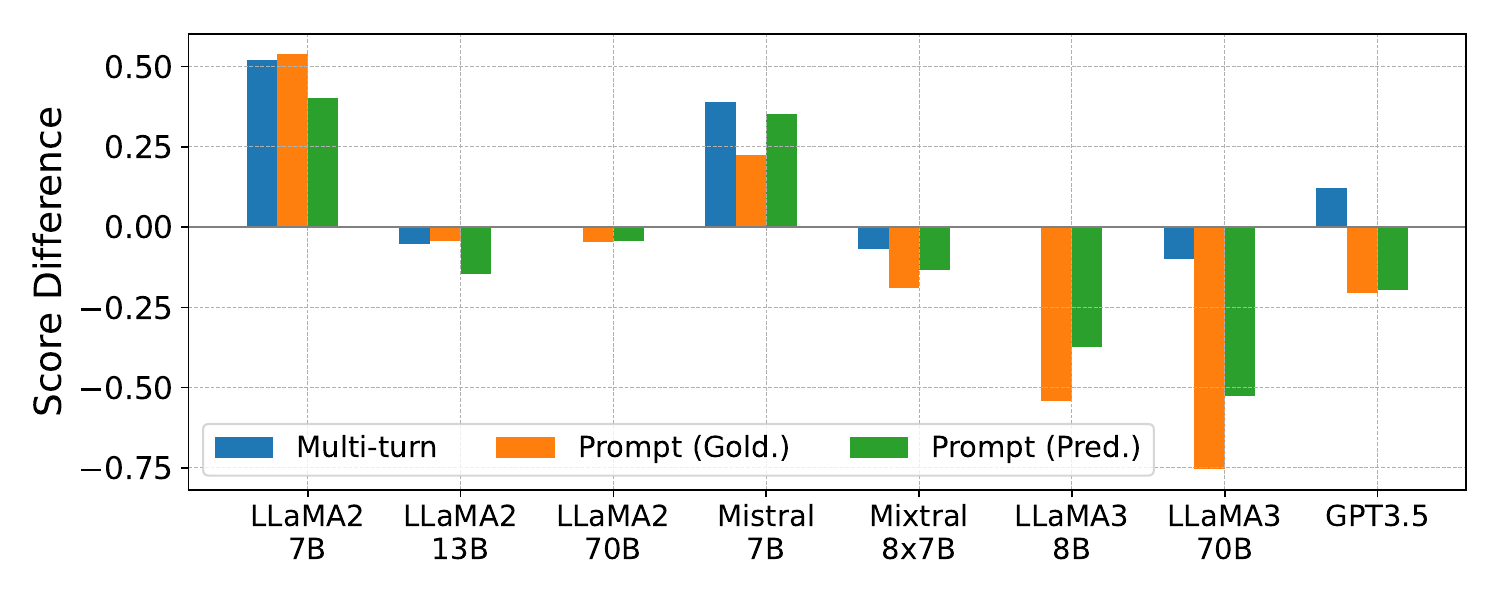}}\\\vspace{-0.2cm}
     \subfloat[Depth 3]{\includegraphics[width=0.43\textwidth]{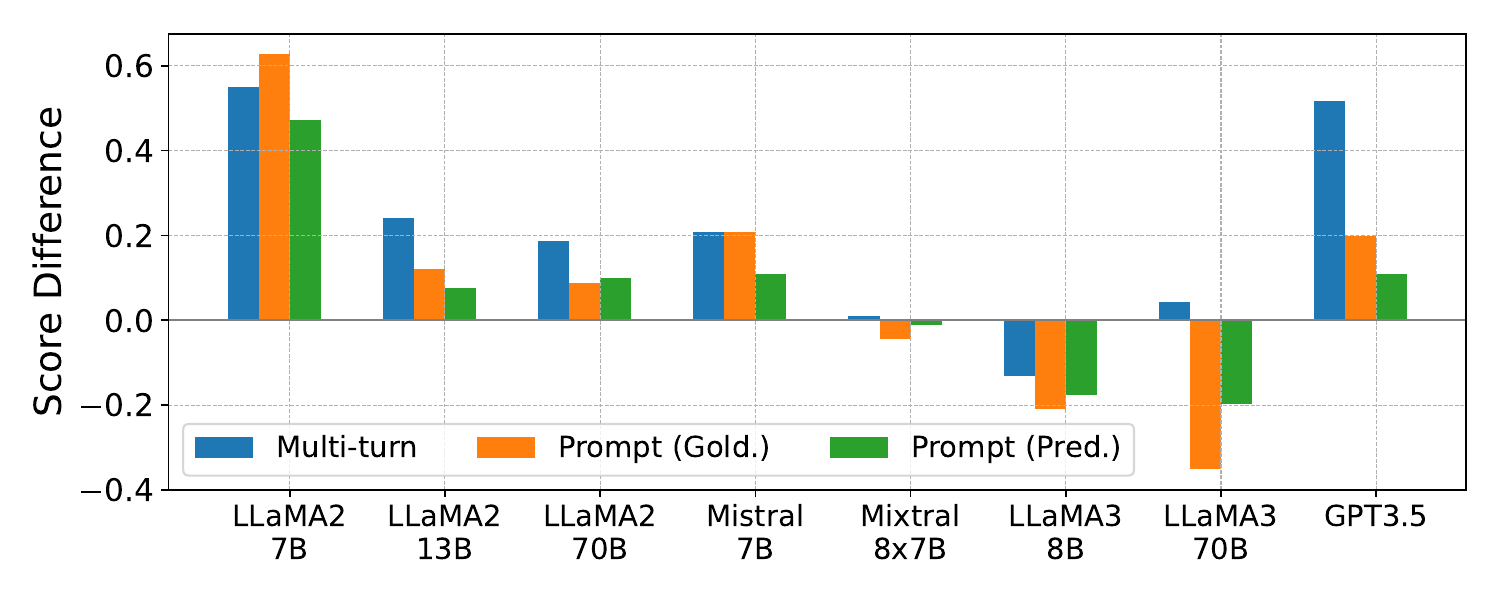}}
    \caption{
     Performance change after providing shallower questions. 
     Note that $D_1$ is not reported for prompt inputs, as $D_1$ does not have shallower questions.
    }
    \vspace{-0.1cm}
    \label{fig:reasoning_input}
    \vspace{-0.3cm}
\end{figure}
\subsection{Effect of Explicit Reasoning Process}\label{subsec:effect_enforcing}
In this study, as presented in Figure~\ref{fig:overview} (a), $D_3$ questions can be solved through sequential reasoning, utilizing answers from $D_1$ to $D_3$ questions. 
Previous studies on implicit reasoning \cite{Wei2022ChainOT, press-etal-2023-measuring, zhou2023leasttomost} have shown that enforcing LLMs to reason through intermediate steps explicitly can improve their reasoning ability. 
We investigate whether explicitly providing these reasoning processes to the model can aid in solving complex questions.

We encourage the model to reason by providing shallower questions in three ways: (i) \textbf{Multi-turn}, where shallower questions are provided as user queries in a multi-turn conversation; (ii) \textbf{Prompt (Gold)}, where shallower questions and their gold answers are provided in prompts; (iii) \textbf{Prompt (Pred.)}, where shallower questions with the model's predictions are given in prompts. 
Note that prompt-based approaches require shallower QA pairs as inputs, which cannot be applied to $D_1$ questions. 
The prompt template for each approach is provided in Appendix~\ref{appendix:inference_prompts}.

\paragraph{Explicitly providing shallower solutions is beneficial for small models and complex questions.}
Figure~\ref{fig:reasoning_input} illustrates the depthwise performance changes after incorporating deconstructed question information. 
Providing shallower questions benefits models with smaller capacities, such as LLaMA 2 7B Chat and Mistral 7B Instruct v0.2. 
For relatively simpler questions ($D_2$), the benefit is less pronounced or may even decrease the performance of more capable models (>7B). However, intermediate questions ($D_2$) are beneficial for complex questions ($D_3$), except for models with large capacities ($\geq$ 56B). 
These findings align with recent research on decomposing a complex question into simpler sub-tasks and solving sub-tasks prior to the final answer~\cite{juneja-etal-2023-small, khot2023decomposed}, which have shown high performance improvements for solving complex problems across different model sizes.

\paragraph{Implicitly guiding reasoning via multi-turn interactions best improves performance.}
When comparing the two prompt-based inputs, smaller models tend to perform better with gold answers (Gold.), while more capable models favor self-prediction results (Pred.). 
This preference likely arises because more capable models align better with their own generated outputs, which reflect their advanced internal reasoning processes. 
The multi-turn approach provides the most stable results across all depths, enhancing the performance of smaller models while causing minimal performance drops for larger models. 
Additionally, the multi-turn approach improves $D_1$ performance by providing context or domain information as part of the interaction history.

\section{Conclusion}
\label{sec:conclusion}

In this study, we explore the reasoning capabilities of LLMs by deconstructing real-world questions into a graph. 
We introduce \dataset, a set of deconstructed $D_3$ questions mapped into a hierarchical graph, requiring utilization of muliple layers of knowledge in the sequence of $D_1$, $D_2$ to $D_3$.
This hierarchical approach provides a comprehensive assessment of LLM performance by measuring forward and backward discrepancies between simpler and complex questions. 
Our comparative analysis of LLMs with different capacities reveals an inverse relationship between model capacities and discrepancies. 
Memorization analysis suggests that the sources of forward and backward discrepancies in large models stem from different types of failures. 
Lastly, we demonstrate that guiding models from shallower to deeper questions through multi-turn interactions stabilizes performance across the majority of models. 
These findings emphasize the importance of intermediate knowledge extraction in understanding LLM reasoning capabilities.

\section*{Limitations}
\label{sec:limitations}
\paragraph{Small sample size} 
Our dataset, {\dataset}, consists of 91 complex ($D_3$) questions from the TutorEval dataset, along with 1,480 derived shallower ($D_2$, $D_1$) questions. Despite the diversity in reasoning types explored (Section~\ref{subsec:data-analysis}) and the hierarchical structuring of subquestions, the limited number of complex questions and the narrow content scope restrict the generalizability of our findings. The selection of TutorEval as our primary source is based on the challenge of manually developing or even sourcing intricate questions that necessitate advanced reasoning skills; such questions require (1) maintaining real-world relevance, (2) eliciting long-form answers, and (3) having minimal risk of test set contamination. Within TutorEval, complex $D_3$ questions represent only 33.6\% of its 834 questions, which further reduces to 10.9\% when excluding questions that require external knowledge retrieval. We encourage future research to build larger, more diverse datasets to more robustly assess knowledge reasoning capabilities of LLMs.

\paragraph{GPT-4 data generation and evaluation}
All questions except for $D_3$ and reference answers in {\dataset} are generated by GPT-4 Turbo. To ensure the quality of these questions, we have established strict decomposition criteria (Section~\ref{subsec:decomposition-criteria}) and implemented rigorous procedures including detailed instructions, question augmentation, manual rewriting and verification by human annotators (Section~\ref{subsec:data-construction}). The reliability of the answers is supported by findings from \citet{chevalier2024language}, which demonstrate GPT-4’s high accuracy of 92\% on TutorEval problems as assessed by human evaluators. However, there may exist inaccuracies due to unseen errors in the decomposition process or unverified knowledge produced by the model.

Furthermore, we utilize GPT-4 Turbo to assess the correctness of model predictions. Following protocols from previous studies \citep{kim2024prometheus, kim2024prometheus2} which highlight GPT-4’s strong correlation with human judgments on long-form content, we provide detailed instructions and specific scoring rubrics to the evaluator to ensure that the evaluation process aligns closely with our objectives. In addition, we conduct human evaluations and compare with GPT-4 Turbo evaluations, and measure sufficiently high inter-annotator agreement (Appendix~\ref{appendix:reliability_llm_judge}). Still, the evaluation method is subject to bias inherent in LLM judges.

\section*{Acknowledgement}
\label{sec:acknowledgement}
We thank Hyeonbin Hwang, Sohee Yang, and Sungdong Kim for constructive feedback and discussions.
This work was partly supported by KAIST-NAVER Hypercreative AI Center and Institute for Information {\&} communications Technology Promotion (IITP) grant funded by the Korea government (MSIT) (RS-2024-00398115, Research on the reliability and coherence of outcomes produced by Generative AI, 30\%).



\bibliography{acl,custom}

\clearpage
\newpage

\appendix

\section{Details in Dataset Construction}\label{appendix:dataset}
\paragraph{Classifying questions based on depth of knowledge}
To categorize questions from the TutorEval dataset~\citep{chevalier2024language}, we use GPT-4 Turbo set at a temperature of 0.7, following the specific prompt detailed in Table~\ref{tab:appendix_classify_questions}. We evaluate the model's classification accuracy using a validation set of 50 questions, which we have previously annotated with their respective depth of knowledge levels. Our optimal prompting strategy involves incorporating key points from each question provided in the original dataset and instructing the model to provide a step-by-step explanation of its classification reasoning. This approach achieves a precision of 0.67 and a recall of 0.77, with a low rate of false positives. Analysis of the entire set of 834 questions reveals the distribution of depth levels: 43\% at $D_2$, 33.6\% at $D_3$, 23.3\% at $D_1$, and only one question at $D_4$. 

\paragraph{$D_3$ question filtering and disambiguation}
From the 280 $D_3$ questions initially identified, we manually exclude questions that are not self-contained, meaning they refer to specific contexts or excerpts in textbook passages that cannot be seamlessly integrated into our input. Examples include questions like, ``I don't understand the point of \textit{Theorems 4.3.2 and 4.3.3.} Why do we care about these statements?'' and ``Please tell me the common conceptual points between \textit{the Weinrich and Wise 1928 study} and \textit{the Roland et al. 1980 paper}.'' Additionally, we disambiguate questions to ensure clarity and context accuracy. For example, the question ``Why is branching unstructured? And is it a bad design choice?'' was initially vague about its reference to `branching.' Upon review, we identify the context as computer programming rather than database systems and revise the question to: ``In the context of computer programming, why is branching considered unstructured, and is it considered a poor design choice?''.

\begin{table}[t]
\centering
\small
\resizebox{0.5\textwidth}{!}{
    \begin{tabularx}{0.5\textwidth}{X}
    \toprule
    \textit{Top-1 before deduplication (similarity = 0.97)}\\
    $D_2$: How do you calculate the determinant of a matrix?\\
    $D_1$: How do you find the determinant of a matrix? \\
    \midrule
    \textit{Top-1 after deduplication (similarity = 0.93)}\\
    $D_2$: What does it mean for two vectors to be orthogonal, and how can you verify this property?\\
    $D_1$: What does it mean for two vectors to be orthogonal? \\
    \bottomrule 
    \end{tabularx}
}
\caption{Top-1 similar question pairs between $D_2$ and $D_1$ before and after the deduplication and augmentation process. While the pair above shares essentially the same depth of knowledge, the pair below substantially differ in knowledge depth due to the $D_2$ question asking additional procedures.}
\label{tab:appendix_dedupe}
\end{table}

\begin{table}[t]
\centering
\small
\resizebox{0.5\textwidth}{!}{
    \begin{tabularx}{0.5\textwidth}{X}
    \toprule
    Describe how division and remainders work when considering congruence modulo a number.\\
    \midrule
    \begin{enumerate}[leftmargin=1.3em, topsep=0pt]
        \setlength\itemsep{-0.15em}
        \item What is the result of a division called?
        \item How is a remainder defined in division?
        \item What does it mean for two numbers to be congruent modulo a number?
        \item \textcolor{red}{What does the term `congruence modulo a number' mean?}
        $\Rightarrow$ \textcolor{blue}{What is the modulo operation in mathematics?}
    \end{enumerate}\\
    \bottomrule 
    \end{tabularx}
}
\caption{The original 4th shallower question (\textcolor{red}{red}) is asking redundant knowledge addressed in the 3rd question. We remove the duplicate question and replace it with a question asking a different concept (\textcolor{blue}{blue}).}
\label{tab:appendix_augment}
\end{table}

\paragraph{Question deduplication and augmentation}
As explained in Section~\ref{subsec:data-construction}, we leverage cosine similarity of question embeddings produced by a Sentence Transformers embedding model\footnote{\href{https://huggingface.co/sentence-transformers/all-mpnet-base-v2}{\texttt{sentence-transformers/all-mpnet-base-v2}}}~\citep{reimers-gurevych-2019-sentence} to identify near-duplicate questions. Specifically, within the same depth 1 or 2, we apply a similarity threshold of 0.9 to identify duplicates and eliminate them. For questions across $D_1$ and $D_2$, we remove $D_2$ questions with a similarity score $\geq 0.9$ to $D_1$ questions and remove $D_1$ questions with a similarity score between 0.8 and 0.9 to $D_1$ questions, under the rationale that the $D_2$ question is effectively at $D_1$ level and vice versa. The similarity threshold and the candidate to remove was determined through qualitative analysis. We analyze that the false positive rate of $D_1$ questions misclassified as $D_2$ decreases from 60\% to 7\% (reduction rate of 88\%). Table~\ref{tab:appendix_dedupe} compares top-1 similar question pairs between $D_2$ and $D_1$ before and after deduplication. We further use GPT-4 Turbo with temperature 0.7 (prompt in Table~\ref{tab:appendix_question_aug_d2} and \ref{tab:appendix_question_aug_d1}) to augment questions filling the gap of knowledge. Table~\ref{tab:appendix_augment} illustrates how an augmented question successfully replaces a previously redundant question, enhancing comprehensiveness.

\begin{table*}
\centering
\small
\begin{tabular}{p{0.475\textwidth}p{0.475\textwidth}}
    \toprule
    \textit{Original question:} & \textit{Flipped question:}\\
    Being congruent modulo an ideal sounds a bit like two numbers being congruent modulo another number. \textbf{Is there a connection} between these two notions? $\Rightarrow$ \textcolor{blue}{Yes} & 
    Are the concepts of being congruent modulo an ideal and two numbers being congruent modulo another number \textbf{entirely distinct with no connection}? $\Rightarrow$ \textcolor{red}{No} \\
    \midrule
    \textit{Predicted answer:} & \textit{Predicted answer:}\\
    \textcolor{blue}{Yes}, there is a connection between the two notions. In mathematics, when we say that two integers are congruent modulo n, it means that they have the same remainder when divided by n. For example, 3 and 6 are congruent modulo 3 because when we divide 3 by 3, the remainder is 0, and when we divide 6 by 3, the remainder is also 0. \lbrack ... \rbrack &
    \textcolor{blue}{Yes}, the concepts of being congruent modulo an ideal and two numbers being congruent modulo another number are distinct. When we say that two numbers a and b are congruent modulo n, we mean that their difference a - b is a multiple of n. This is written as a $\equiv$ b (mod n). \lbrack ... \rbrack \\
    \bottomrule
\end{tabular}
\caption{Example of Mistral 7B Instruct v0.2 favoring affirmative responses over negative responses when the knowledge required is consistent but only the question format is flipped.}
\label{tab:appendix_bias}
\end{table*}

\begin{table}[t]
\centering
\small
\resizebox{0.5\textwidth}{!}{
    \begin{tabularx}{0.5\textwidth}{X}
    \toprule
    Are there problems that one can use standard induction to prove but cannot use strong induction to prove?\\
    $\Rightarrow$ What kind of problems can be proven using standard induction but not strong induction? \\
    \midrule
    If I understand correctly, adding sine functions always results in a new sine function?\\
    $\Rightarrow$ Clarify my understanding that adding sine functions always results in a new sine function. \\
    \midrule
    Can a linear transformation map all points of a vector space to a single point, and under what conditions does this occur?\\
    $\Rightarrow$ Describe the possibility of a linear transformation mapping all points of a vector space to a single point. Under what conditions does this occur? \\
    \bottomrule 
    \end{tabularx}
}
\caption{Example conversions of a binary question into a non-binary question.}
\label{tab:appendix_debias}
\end{table}

\paragraph{Motivation of question debiasing}
In our preliminary study, we found that models tend to favor ``yes'' over ``no'' at the beginning of the response to a question that can be answered in binary format, as exemplified in Table~\ref{tab:appendix_bias}. We recognize that the inherent positivity bias in models~\citep{augustine2011positivity, dodds2015human, lesswrong2023biased} has the potential to skew the model’s reasoning processes and consequently obscure a true evaluation of its capability to reason and articulate nuanced thoughts. To mitigate this, we debias problematic questions by reframing them into more exploratory inquiries.  Example transformations are in Table~\ref{tab:appendix_debias}.

\section{Human Verification on Data Quality}\label{appendix:data_quality_check}

2 of the authors and one graduate student who volunteered annotate 27.5\% of {\dataset}, verifying the three criteria we hold in Section~\ref{subsec:decomposition-criteria}: Comprehensiveness (C1), Implicitness (C2), and Non-binary questioning (C3).  Comprehensiveness and Implicitness are especially crucial criteria for sub-questions to ensure the hierarchy in the reasoning process, as Comprehensiveness ensures no critical knowledge gaps with increasing depth, while Implicitness ensures no straightforward clues, encouraging implicit reasoning between sub-questions.

\begin{table}[ht]
\centering
\small
\begin{tabular}{lrrrr}
\toprule
\multirow{2}{*}{\textbf{C1. Comprehensiveness}} & \multicolumn{2}{c}{$D_3$ $\rightarrow$ $D_2$} & \multicolumn{2}{c}{$D_2$ $\rightarrow$ $D_1$} \\
\cmidrule(lr){2-3} \cmidrule(lr){4-5}
 & Count & \% & Count & \% \\
\midrule
Comprehensive & 22 & 88.0 & 79 & 79.0 \\
Partially comprehensive & 3 & 12.0 & 18 & 18.0 \\
Insufficient & 0 & 0.0 & 3 & 3.0 \\
\bottomrule
\end{tabular}
\caption{Human annotation on Comprehensiveness of a subset of {\dataset} question relations.}
\label{tab:comprehensiveness_check}
\end{table}

\begin{table}[ht]
\centering
\small
\begin{tabular}{lrrrr}
\toprule
\multirow{2}{*}{\textbf{C2. Implicitness}} & \multicolumn{2}{c}{$D_2$} & \multicolumn{2}{c}{$D_1$} \\
\cmidrule(lr){2-3} \cmidrule(lr){4-5}
 & Count & \% & Count & \% \\
\midrule
Fully implicit & 87 & 87.0 & 364 & 91.9 \\
Partially comprehensive & 13 & 13.0 & 31 & 7.8 \\
Insufficient & 0 & 0.0 & 1 & 0.3 \\
\bottomrule
\end{tabular}
\caption{Human annotation on Implicitness of a subset of {\dataset} sub-questions.}
\label{tab:implicitness_check}
\end{table}

\begin{table}[ht]
\centering
\small
\resizebox{\linewidth}{!}{
    \begin{tabular}{lrrrrrr}
    \toprule
    \multirow{2}{*}{\textbf{\shortstack{C3. Non-binary\\Questioning}}} & \multicolumn{2}{c}{$D_3$} & \multicolumn{2}{c}{$D_2$} & \multicolumn{2}{c}{$D_1$}\\
    \cmidrule(lr){2-3} \cmidrule(lr){4-5} \cmidrule(lr){6-7}
     & Count & \% & Count & \% & Count & \%\\
    \midrule
    Open-ended & 24 & 96.0 & 100 & 100.0 & 396 & 100.0\\
    Binary & 1 & 4.0 & 0 & 0.0 & 0 & 0.0\\
    \bottomrule
    \end{tabular}
}
\caption{Human annotation on Non-binary questioning of a subset of {\dataset} sub-questions.}
\label{tab:non_binary_check}
\end{table}

\begin{figure*} 
    \centering
    \includegraphics[width=1\linewidth]{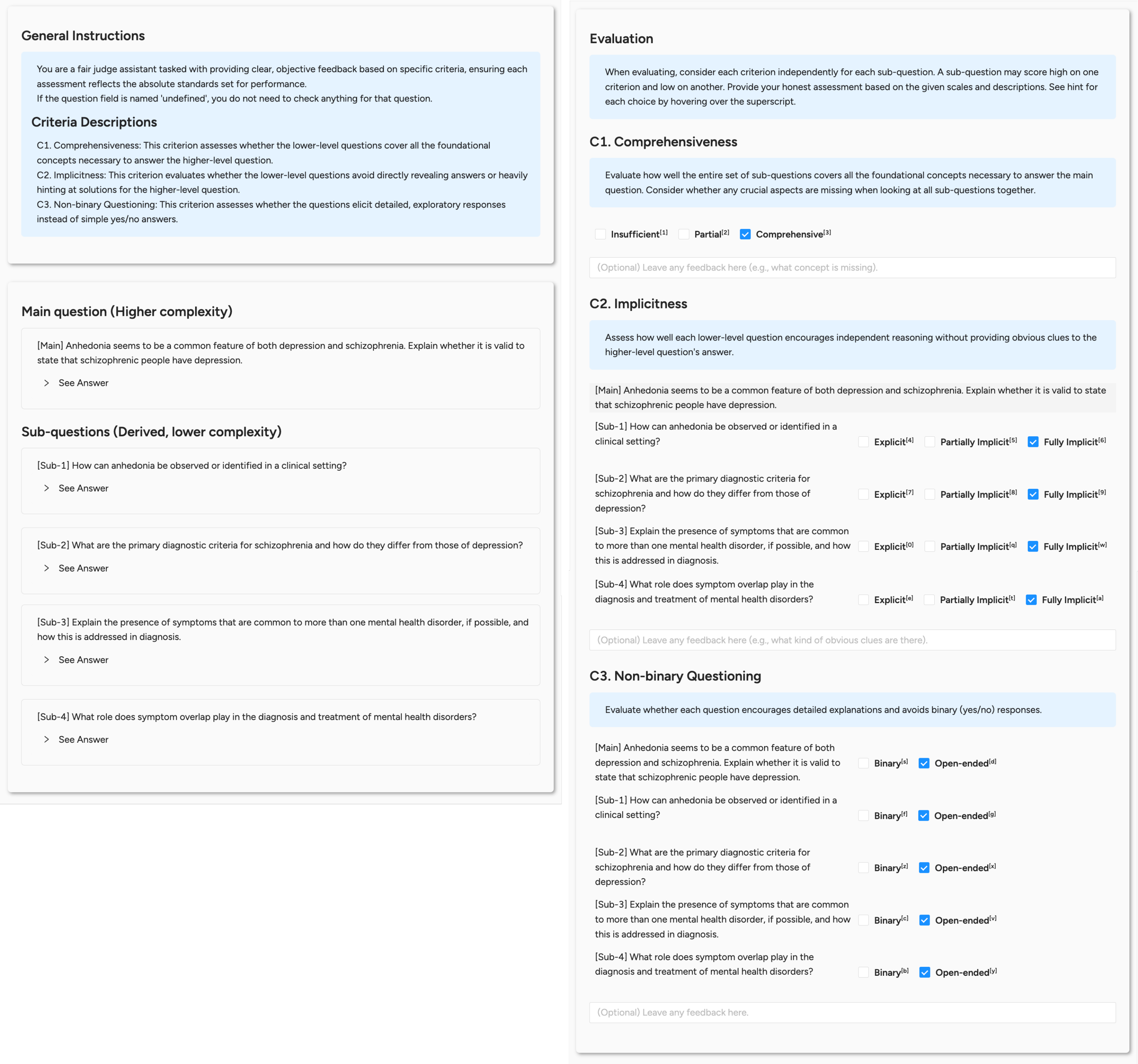}
    \caption{Interface for human annotators to check if Comprehensiveness (C1), Implicitness (C2), Non-binary questioning (C3) hold between a question and its sub-questions in {\dataset}.}
    \label{fig:interface_data_quality}
\end{figure*}

To set up the procedure, we randomly sample 5 $D_3$ questions from each of the 5 domains in our dataset and use all questions derived from the selected $D_3$ questions, totaling 25 $D_3$, 100 $D_2$, and 396 $D_1$ questions. Given 25 $D_3 \rightarrow D_2$ and 100 $D_2 \rightarrow D_1$ relations, the relations are divided into 40, 40, 45 and are assigned to the three workers. For each relation, the main question and the sub-questions (predecessors) are provided along with their gold answers. Then the labeler is asked to check whether the relation is conceptually comprehensive and whether each question is implicit or non-binary. The labeler can choose from three varying degrees of comprehensiveness and implicitness due to the subjective nature of the criteria. The annotation interface is shown in Figure~\ref{fig:interface_data_quality}.

Table~\ref{tab:comprehensiveness_check}, \ref{tab:implicitness_check}, and \ref{tab:non_binary_check} reports the annotation statistics. Table~\ref{tab:non_binary_check} shows that the decompositions into shallower questions are fully comprehensive (C1) in 88\% of $D_3 \rightarrow D_2$ relations and 79\% of $D_2 \rightarrow D_1$ relations, reaching 100\% and 97\% when taking partially comprehensive relations as well, respectively. Also, Table~\ref{tab:implicitness_check} shows that 87\% of $D_2$ and 91.9\% of $D_1$ questions do not hint at solutions for more complex questions (C2), with similarly low failure rates. We also find in Table~\ref{tab:non_binary_check} that nearly all questions require open-ended answers (C3). Human verification data provides evidence that our synthetically generated edges in the adequately represent the reasoning process.

\section{Dataset License} 
The TutorEval~\citep{chevalier2024language} dataset from which we source complex questions has not disclosed the license yet. Our {\dataset} is subject to OpenAI's Terms of Use for the generated data. We will notify the intended use of our dataset for research when releasing our dataset to the public. 

\section{Reasoning Type Analysis}\label{appendix:analysis}

\begin{table}[t]
\centering
\small
\label{tab:reasoning_distribution}
\begin{tabular}{lrrrr}
\toprule
\multirow{2}{*}{\textbf{Reasoning Type}} & \multicolumn{2}{c}{\textbf{Depth 3}} & \multicolumn{2}{c}{\textbf{Depth 2}} \\
\cmidrule(lr){2-3} \cmidrule(lr){4-5}
& Count & \%     & Count & \%     \\
\midrule
Comparative             & 12    & 21.1 & 19    & 11.6 \\
Relational              & 10    & 17.5 & 37    & 22.6 \\
Causal                  & 6     & 10.5 & 19    & 11.6 \\
Inductive               & 5     & 8.8  & 6     & 3.7  \\
Criteria Development    & 5     & 8.8  & 13    & 7.9  \\
Procedural              & 4     & 7.0  & 22    & 13.4 \\
Evaluative              & 4     & 7.0  & 12    & 7.3  \\
Example                 & 2     & 3.5  & 8     & 4.9  \\
Quantitative            & 2     & 3.5  & 6     & 3.7  \\
Application             & 2     & 3.5  & 19    & 11.6 \\
Other                   & 5     & 8.8  & 3     & 1.8  \\
\midrule
Total                   & 57    & 100  & 164   & 100 \\
\bottomrule
\end{tabular}
\caption{Distribution of reasoning types for $D_3$ and $D_2$ in a subset of {\dataset}. Multiple reasoning types can be included in one instance.}
\label{tab:appendix_reasoning_type}
\end{table}

In Table~\ref{tab:appendix_reasoning_type}, we report the distribution of reasoning types annotated by the authors on a sample of 20 $D_3$ questions and $D_2$ and $D_2$ related to them. Table~\ref{tab:appendix_all_types} outlines the definition of each reasoning type and a representative example set of questions that best elicits such reasoning.  We provide question deconstructions examples in Table~\ref{tab:appendix_eigenvector_example} and Table~\ref{tab:appendix_mule_example} where each showcases distinct reasoning types and knowledge.

\section{Details in Main Experiments}\label{appendix:experiments}
\subsection{Model Inference}
To inference LLMs used in our experimental setup (Section~\ref{subsec:setup}), we use a standardized API from OpenRouter\footnote{\href{https://openrouter.ai/}{\texttt{openrouter.ai}}} to access LLMs and use the complementary LiteLLM\footnote{\href{https://litellm.vercel.app/docs/providers/openrouter}{\texttt{litellm.vercel.app/docs/providers/openrouter}}} interface to call model generations. An exception is LLaMA 7B Chat, which is not hosted in OpenRouter; we use the HuggingFace model and the vLLM~\citep{kwon2023efficient} inference engine for this particular model, performing local inference with mixed precision on 1 NVIDIA A6000 40GB GPU. We use the default sampling parameters suited for each model. The specific prompt templates used to induce reasoning paths are organized in Appendix~\ref{appendix:inference_prompts}.
The inference on the whole pass of {\dataset} finishes within 10 minutes. We report single-run results.

\begin{table*}[ht]\centering
\scriptsize
\begin{tabular}{lccccccccc}\toprule
\multirow{2}{*}{\textbf{Model}} &\multicolumn{3}{c}{\textbf{Average Forward Discrepancy}} &\multicolumn{3}{c}{\textbf{Value}} &\multicolumn{3}{c}{\textbf{Frequency (\%)}} \\\cmidrule(lr){2-4}\cmidrule(lr){5-7}\cmidrule(lr){8-10}
&$D_2 \rightarrow D_3$ &$D_1 \rightarrow D_2$ & Overall&$D_2 \rightarrow D_3$ &$D_1 \rightarrow D_2$ & Overall&$D_2 \rightarrow D_3$ &$D_1 \rightarrow D_2$ & Overall \\\toprule
LLaMA 2 7B Chat &0.1304 &0.1814 &0.1756 &0.2708 &0.2683 &0.2685 &48.15 &67.62 &65.40 \\
LLaMA 2 13B Chat &0.1524 &0.1582 &0.1573 &0.2572 &0.2720 &0.2697 &59.26 &58.14 &58.31 \\
LLAMA 2 70B Chat &0.1259 &0.1361 &0.1344 &0.2633 &0.2490 &0.2512 &47.83 &54.68 &53.50 \\
Mistral 7B Instruct v0.2 &0.0920 &0.1569 &0.1474 &0.2031 &0.2294 &0.2267 &45.28 &68.39 &65.01 \\
Mixtral 8x7B Instruct v0.1&0.0868 &0.0791 &0.0806 &0.1844 &0.2058 &0.2009 &47.06 &38.46 &40.14 \\
Llama 3 8B Instruct &0.0831 &0.0957 &0.0934 &0.2225 &0.2258 &0.2253 &37.33 &42.38 &41.44 \\
Llama3 70B Instruct &0.0653 &0.0497 &0.0528 &0.2176 &0.2211 &0.2202 &30.00 &22.47 &23.99 \\
GPT-3.5 Turbo &0.1002 &0.0722 &0.0779 &0.1608 &0.1369 &0.1424 &62.35 &52.73 &54.70 \\
\bottomrule
\end{tabular}
\caption{Average intensity and frequency of \fwd.}\label{tab:appendix_fwd}
\end{table*}

\begin{table*}[ht]\centering
\scriptsize
\begin{tabular}{lccccccccc}\toprule
\multirow{2}{*}{\textbf{Model}} &\multicolumn{3}{c}{\textbf{Average Backward Discrepancy}} &\multicolumn{3}{c}{\textbf{Value}} &\multicolumn{3}{c}{\textbf{Frequency (\%)}} \\\cmidrule(lr){2-4}\cmidrule(lr){5-7}\cmidrule(lr){8-10}
&$D_2 \rightarrow D_3$ &$D_1 \rightarrow D_2$ & Overall&$D_2 \rightarrow D_3$ &$D_1 \rightarrow D_2$ & Overall&$D_2 \rightarrow D_3$ &$D_1 \rightarrow D_2$ & Overall \\\toprule
LLaMA 2 7B Chat &0.2193 &0.1104 &0.1342 &0.3827 &0.3589 &0.3671 &57.31 &30.77 &36.57 \\
LLaMA 2 13B Chat &0.1255 &0.0782 &0.0879 &0.3846 &0.3339 &0.3473 &32.64 &23.43 &25.32 \\
LLAMA 2 70B Chat &0.1363 &0.0632 &0.0787 &0.3811 &0.3258 &0.3442 &35.76 &19.40 &22.88 \\
Mistral 7B Instruct v0.2 &0.1442 &0.0700 &0.0881 &0.3488 &0.3071 &0.3225 &41.33 &22.81 &27.31 \\
Mixtral 8x7B Instruct v0.1 &0.0627 &0.0635 &0.0633 &0.2979 &0.2728 &0.2781 &21.04 &23.27 &22.76 \\
Llama 3 8B Instruct &0.0878 &0.0717 &0.0752 &0.3500 &0.3141 &0.3227 &25.08 &22.82 &23.32 \\
Llama3 70B Instruct &0.0427 &0.0442 &0.0438 &0.2778 &0.2692 &0.2710 &15.38 &16.41 &16.18 \\
GPT-3.5 Turbo &0.0457 &0.0672 &0.0626 &0.2892 &0.2602 &0.2644 &15.79 &25.81 &23.68 \\
\bottomrule
\end{tabular}
\caption{Average intensity and frequency of \bwd.}\label{tab:appendix_bwd}
\end{table*}

\subsection{LLM-as-a-Judge Evaluation}
When prompting GPT-4 Turbo to evaluate model responses, we use a temperature of 1.0, nucleus sampling with \texttt{top\_p} of 0.9, and maximum number of generation tokens of 1,024, following previous works~\citep{ye2024flask, kim2024prometheus, kim2024prometheus2, lee2024prometheus}. The prompt template including the score rubric is in Table~\ref{tab:appendix_eval}. We report single-run results. See Table~\ref{tab:appendix_qa_example} for example output format. 
Unlike prior works that emphasize the use of instance-specific scoring rubrics~\citep{kim2024prometheus, kim2024prometheus2, lee2024prometheus}, our initial experiments comparing evaluations given a common rubric and instance-specific rubric showed that instance-specific rubrics increase noise in evaluation and decrease the quality of evaluation. We speculate that it is because the focus of our evaluation is on a \textit{common} factor of factual correctness, \textit{i.e.}, whether the model accurately uses knowledge in the reasoning process, different from conventional benchmark evaluations.

\section{Reliability of LLM-as-a-Judge}\label{appendix:reliability_llm_judge}
\begin{table}[t]
\centering
\small
\begin{tabular}{lcc}
\toprule
\textbf{Question depth} & \textbf{Human-Human} & \textbf{Human-GPT-4} \\
\midrule
\multirow{2}{*}{$D_3$} & 0.4848 & 0.7064 \\
   & \scriptsize{(n = 3)} & \scriptsize{(n = 13)} \\[0.5ex]
\multirow{2}{*}{$D_2$} & 0.6464 & 0.7730 \\
   & \scriptsize{(n = 6)} & \scriptsize{(n = 32)} \\[0.5ex]
\multirow{2}{*}{$D_1$} & 0.5671 & 0.7969 \\
   & \scriptsize{(n = 11)} & \scriptsize{(n = 55)} \\[0.5ex]
\midrule
\multirow{2}{*}{Overall} & 0.5730 & 0.7797 \\
        & \scriptsize{(n = 20)} & \scriptsize{(n = 100)} \\
\bottomrule
\end{tabular}
\caption{Krippendorf's Alpha between human-human and human-GPT-4 ratings on model responses to {\dataset} questions. For human-GPT agreement, the scores of predictions rated by the three human raters are averaged. The number of responses in each measurement is reported below the Krippendorf's Alpha value.}
\label{tab:krippendorf}
\end{table}

\begin{figure*}
    \centering
    \includegraphics[width=1\linewidth]{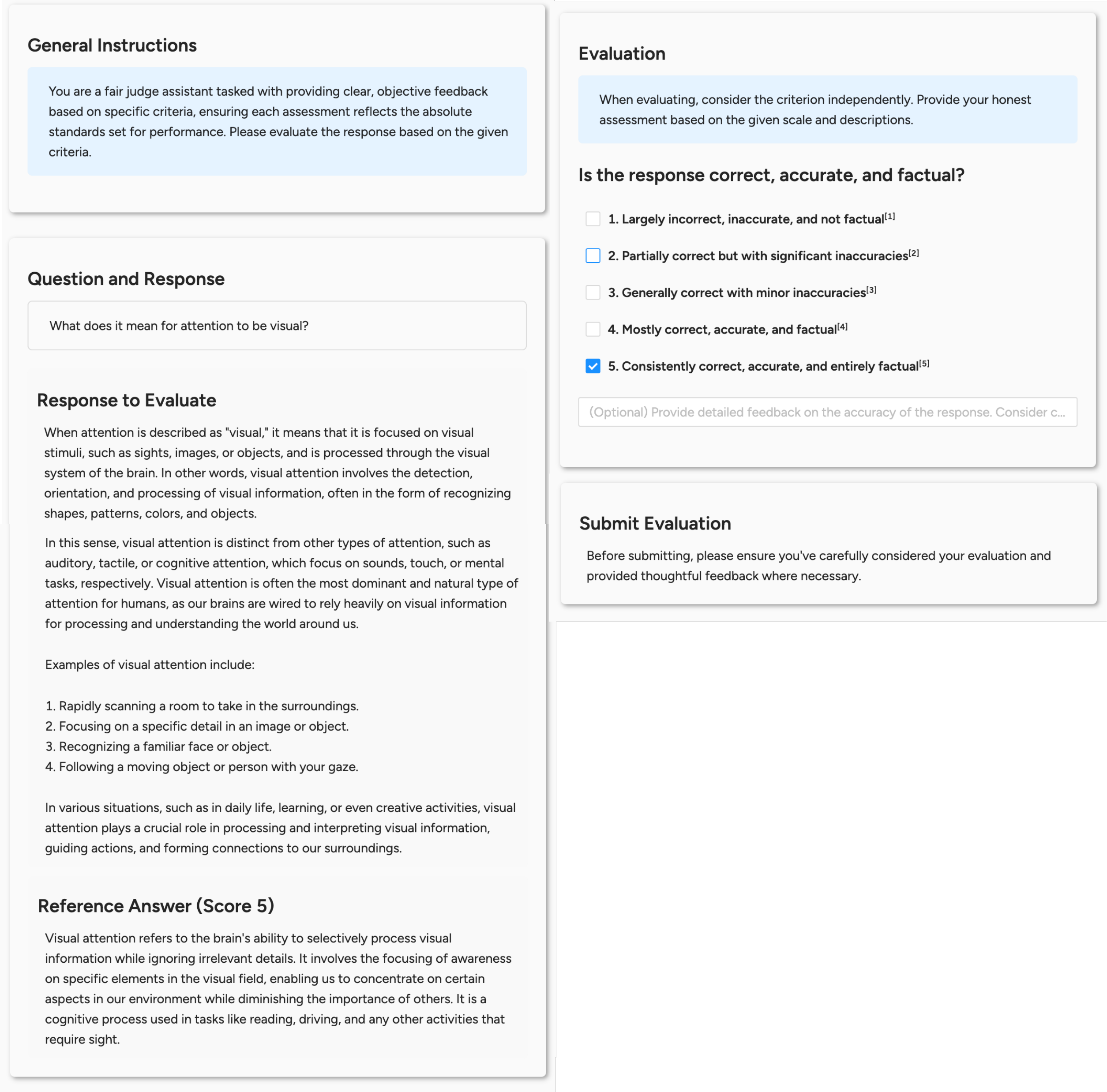}
    \caption{Interface for human evaluators to evaluate an LLM's response on a question from {\dataset}. The rubric shown is a simplified form of the actual factual accuracy rubric used in LLM evaluations.}
    \label{fig:interface_human_eval}
\end{figure*}

To assess the reliability of LLM evaluations in our analysis, we conduct human evaluation of LLM responses and calculate the agreement between annotations. We randomly sample 20 model responses for each score level (1 to 5) as evaluated by GPT-4 Turbo, with the question and response model being random as well. 2 of the authors and one graduate student who volunteered evaluate 46, 46, and 48 unique responses, respectively, and all 3 workers label the remaining 20 responses set aside for inter-annotator agreement. The human raters are given only 1 instance at a time and individually score it on a scale of 1 to 5, under the exact setting of our LLM-as-a-Judge experiments. The evaluation interface is shown in Figure~\ref{fig:interface_human_eval}. Following \citet{ye2024flask}, we measure Krippendorf's Alpha~\citep{krippendorff2018content, castro-2017-fast-krippendorff} with an ordinal metric to reliability between three human raters and between humans and GPT-4 Turbo. 

Table~\ref{tab:krippendorf} reports the agreement results. The results show that the human-GPT agreement is substantially high, approaching 0.80, the commonly accepted reliability threshold~\citep{krippendorff2018content}. While the sample size is smaller, there is also moderate human-human agreement. This implies that the individual absolute rating scheme is effective and that GPT-4 Turbo evaluations are aligned with humans in our setting.


\section{Discrepancy Results}\label{appendix:discrepancy}
To separately observe how frequently each discrepancy occurs and its intensity when it happens, Table~\ref{tab:appendix_fwd} and Table~\ref{tab:appendix_bwd} show the average intensity and frequency of each forward and backward discrepancy. Note that the average discrepancy is calculated as the product of the value and frequency. Overall, forward discrepancies appeared more frequently, although their intensity was relatively low (between 0.14 and 0.26). 
In contrast, backward discrepancies appeared less than 25\%, except for LLaMA 2 7B, which exhibited high intensity (between 0.26 and 0.37).

\begin{figure*}[ht]
    \centering
    \subfloat[LLaMA 2 13B Chat]{\includegraphics[width=0.3\textwidth]{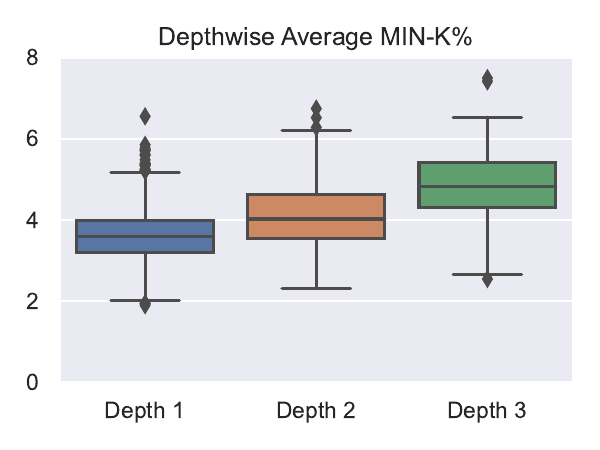}}
    \subfloat[Mistral 8B Instruct]{\includegraphics[width=0.3\textwidth]{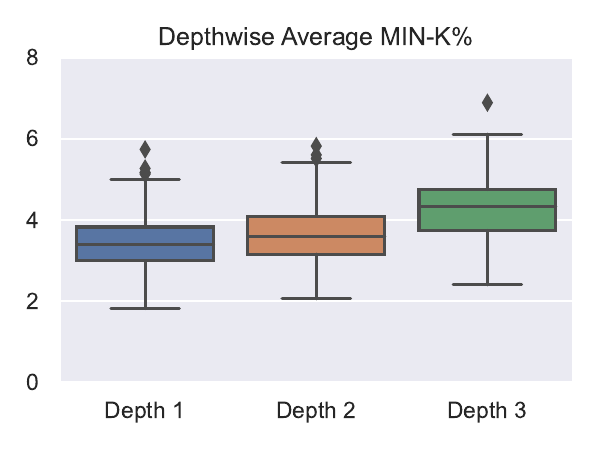}}
    \subfloat[Mixtral 8x7B Instruct]{\includegraphics[width=0.3\textwidth]{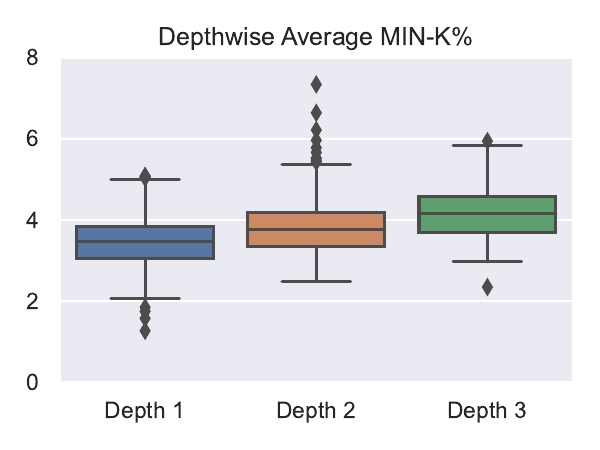}}
    \caption{Average Min-K\% probability at each depth. Lower values indicate more memorization while higher values indicate less memorization.}
    \label{fig:appendix_min_k}
\end{figure*}

\section{Overall Results with Min-K\% Probability}\label{appendix:min_k_prob}
\subsection{Depthwise Min-K\% Prob.}
In Figure~\ref{fig:appendix_min_k}, we plot the Min-k\% probability of LLaMA 2 13B Chat, Mistral 8B Instruct and Mixtral 8x7B Instruct.  
Similar to Figure~\ref{fig:memorization}, $D_3$ shows the highest average Min-K\% probability, indicating the least memorization over all three models.

\subsection{Score Gap within Neighboring Questions}
\begin{figure*}[ht]
    \centering
    \subfloat[LLaMA 2 13B Chat]{\includegraphics[width=0.4\textwidth]{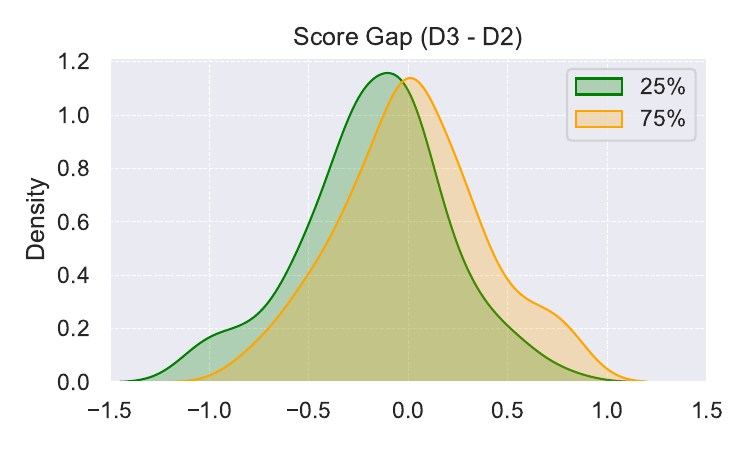}}
    \subfloat[LLaMA 3 8B Chat]{\includegraphics[width=0.4\textwidth]{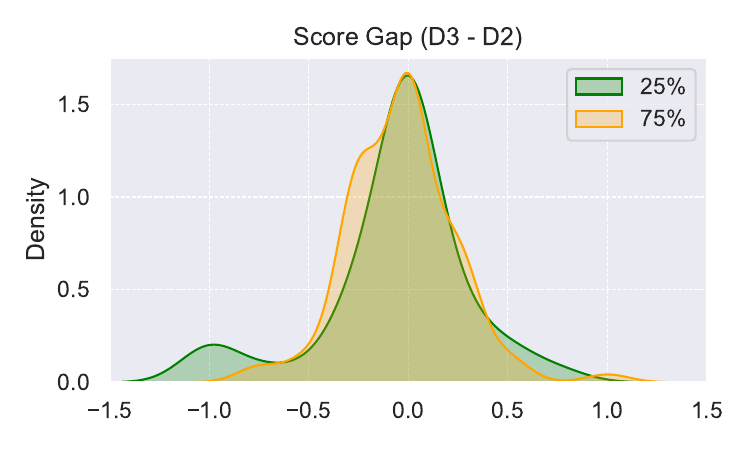}}
    \vspace{-0.1cm}
    \subfloat[Mixtral 8x7B Instruct]{\includegraphics[width=0.4\textwidth]{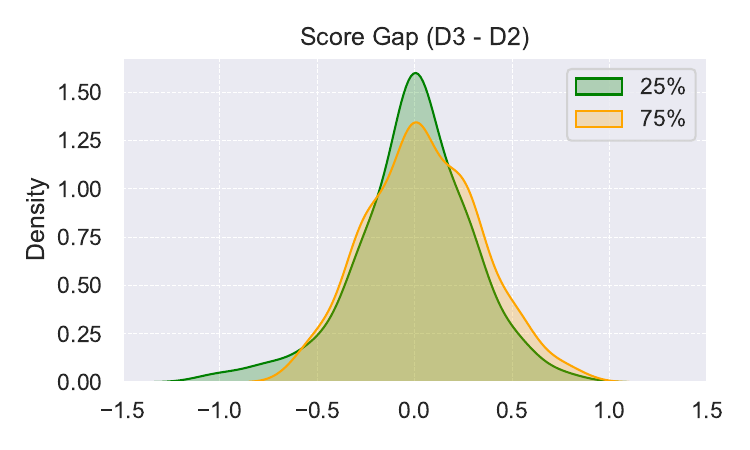}}
    \subfloat[Mixtral 8x7B Instruct]{\includegraphics[width=0.4\textwidth]{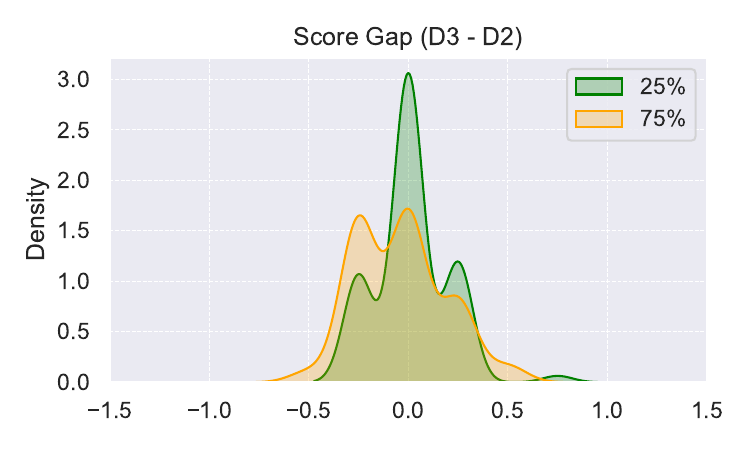}}
    \caption{Factual accuracy difference between neighboring $q_3$ and $q_2$ in bottom 25\% and top 75\% quantiles. Positive gap indicates \bwd~and negative gap represents \fwd.}
    \label{fig:appendix_score_gap}
\end{figure*}

Figure~\ref{fig:appendix_score_gap} presents the KDE plot of the factual accuracy gap between $q_3$ and $q_2$ for $q_3$ instances whose Min-\%K probability is in the bottom 25\% and top 75\%. A positive gap represents higher factual accuracy for $q_3$, indicating \bwd. In contrast, a negative difference represents \fwd.

\begin{table*}[ht]\centering
\scriptsize
\begin{tabular}{p{1cm}p{3.8cm}p{9.5cm}}\toprule
\textbf{Reasoning Type} &\textbf{Explanation} &\textbf{Example} \\\midrule
Comparative &Compare two or more concepts, identifying similarities and differences. &{\two} How do neutrinos differ from other subatomic particles, and why are they considered potential candidates for dark matter?\newline
\hspace*{4mm}{\one} What are neutrinos?\newline
\hspace*{4mm}{\one} What are subatomic particles?\newline
\hspace*{4mm}{\one} What is dark matter?\newline
\hspace*{4mm}{\one} What characteristics do particles need to be considered candidates for dark matter?\\\midrule
Relational &Specify and explain relationships and understand how different concepts are connected organically. &{\two} Describe how eco-efficient urban planning can address the challenges of rapid urbanization in developing countries.\newline
\hspace*{4mm}{\one} What is urbanization?\newline
\hspace*{4mm}{\one} What does eco-efficient mean?\newline
\hspace*{4mm}{\one} What are common challenges faced by rapidly urbanizing cities in developing countries?\newline
\hspace*{4mm}{\one} What is urban planning?\\\midrule
Causal &Identify cause-and-effect relationships. &{\two} Explain how the bending of stereocilia on hair cells leads to the depolarization of these cells.\newline
\hspace*{4mm}{\one} What are stereocilia?\newline
\hspace*{4mm}{\one} What is depolarization?\newline
\hspace*{4mm}{\one} Where are hair cells located?\newline
\hspace*{4mm}{\one} What is the function of hair cells in the ear?\\\midrule
Inductive &Make broad generalizations from specific observations and/or formulate a hypothesis about a particular concept. &{\three} Can you sum up the point of connecting finite sums to integrals? This concept is still a bit obscure to me.\newline
\hspace*{4mm}{\two} How do you approximate the area under a curve using rectangles or trapezoids?\newline
\hspace*{4mm}{\two} Explain the process of taking the limit of a sum as the number of rectangles increases to infinity.\newline
\hspace*{4mm}{\two} What is a Riemann sum, and how is it related to the concept of an integral?\newline
\hspace*{4mm}{\two} How can finite sums be used to estimate real-world quantities that change continuously over an interval?\\\midrule
Criteria Development &Understand when and why specific criteria apply, and know the conditions or assumptions required for different cases. &{\two} Under what conditions does the ideal gas law provide accurate predictions, and when does it not?\newline
\hspace*{4mm}{\one} What is the ideal gas law equation?\newline
\hspace*{4mm}{\one} What are the standard conditions for temperature and pressure in experiments?\newline
\hspace*{4mm}{\one} What is meant by 'ideal gas'?\newline
\hspace*{4mm}{\one} How do real gases differ from ideal gases?\\\midrule
Procedural &Select a procedure according to task need and perform it. &{\two} Describe the process by which hair cells transduce mechanical energy from sound waves into electrical signals.\newline
\hspace*{4mm}{\one} What are hair cells?\newline
\hspace*{4mm}{\one} What is mechanical energy?\newline
\hspace*{4mm}{\one} What are sound waves?\newline
\hspace*{4mm}{\one} What are electrical signals?\\\midrule
Evaluative &Verify reasonableness of results. &{\three} How can I evaluate the suitability of the ideal gas equation for a given gas?\newline
\hspace*{4mm}{\two} How do you calculate the properties such as pressure, temperature, and volume using the ideal gas law?\newline
\hspace*{4mm}{\two} What methods can be used to obtain experimental data for gas properties under specific conditions?\newline
\hspace*{4mm}{\two} How can deviations from ideal gas behavior be identified and measured?\newline
\hspace*{4mm}{\two} Under what conditions does the ideal gas law provide accurate predictions, and when does it not?\\\midrule
Example &Provide example for the given concept. &{\two} Describe a scenario where energy is conserved but the process is thermodynamically impossible.\newline
\hspace*{4mm}{\one} What does the law of conservation of energy state?\newline
\hspace*{4mm}{\one} What is thermodynamic impossibility?\newline
\hspace*{4mm}{\one} What is meant by energy conversion?\newline
\hspace*{4mm}{\one} Can energy be created or destroyed?\\\midrule
Quantitative &Manipulate numerical data to make informed decisions. &{\two} Explain the process and time complexity of deleting an element from a data structure like a linked list.\newline
\hspace*{4mm}{\one} What is a linked list?\newline
\hspace*{4mm}{\one} What is the definition of time complexity?\newline
\hspace*{4mm}{\one} How is data stored in a linked list?\newline
\hspace*{4mm}{\one} What does 'deleting an element' mean in the context of data structures?\\\midrule
Application &Apply concepts to practical situations. &{\two} What policies can governments implement to encourage the transition towards a circular economy and sustainable business practices?\newline
\hspace*{4mm}{\one} What is a circular economy?\newline
\hspace*{4mm}{\one} What are sustainable business practices?\newline
\hspace*{4mm}{\one} What is the role of government in regulating the economy?\newline
\hspace*{4mm}{\one} What does the term 'policy' mean in the context of government regulation?\\
\bottomrule
\end{tabular}
\caption{Reasoning type explanation and examples. $D_3$, $D_2$, and $D_1$ questions are denoted as {\three}, {\two}, {\one}, respectively.}\label{tab:appendix_all_types}
\end{table*}

\begin{table*}[ht]
\centering
\small
\begin{tabularx}{0.7\textwidth}{X}
\toprule
{\three} Does a matrix always have a basis of eigenvectors? \\
\hspace{4mm} {\two} How can you determine if a square matrix is diagonalizable?\\
\hspace{8mm} {\one} What is the definition of a square matrix?\\
\hspace{8mm} {\one} What are the characteristics of a diagonal matrix?\\
\hspace{8mm} {\one} What is meant by the eigenvalues of a matrix?\\
\hspace{8mm} {\one} How is the characteristic equation of a matrix defined?\\
\hspace{4mm} {\two} What is the process for finding the eigenvalues of a matrix?\\
\hspace{4mm} {\two} Explain how to compute eigenvectors from a given set of eigenvalues.\\
\hspace{4mm} {\two} Describe the method to perform a similarity transformation on a matrix. \\
\bottomrule
\end{tabularx}
\caption{Snippet of hierarchical question deconstruction for analyzing matrix diagonalizability. The topmost complex question, $D_3$, requires developing criteria of whether the statement holds or not. The first $D_2$ question identifies the key property to help determine the case. The $D_1$ child questions addresses relevant definitions, characteristics, and formula in order to synthesize the foundational concepts.}\label{tab:appendix_eigenvector_example}
\end{table*}

\begin{table*}[ht]
\centering
\small
\begin{tabularx}{0.9\textwidth}{X}
\toprule
{\three} I thought that animals from different species could not produce viable offspring. However, a horse and a donkey can produce viable offspring. Isn't this contradictory? \\
\hspace{4mm} {\two} How do horses and donkeys differ genetically and what makes them separate species?\\
\hspace{4mm} {\two} What is the definition of a viable offspring in biological terms?\\
\hspace{4mm} {\two} Can you explain the concept of hybrid vigor and how it applies to cross-species offspring such as mules?\\
\hspace{8mm} {\one} What is the definition of hybrid vigor?\\
\hspace{8mm} {\one} What is a mule?\\
\hspace{8mm} {\one} How is a mule produced?\\
\hspace{8mm} {\one} What are the parent species of a mule?\\
\hspace{4mm} {\two} What are the implications of sterility in hybrids like mules for their classification as members of a species? \\
\bottomrule
\end{tabularx}
\caption{Snippet of hierarchical question deconstruction for understanding species differentiation and hybrid viability. Key reasoning steps include identifying exceptions, elucidating causal relationships, and hypothesizing potential outcomes in species classification criteria, culminating in a robust evaluation of biological definitions and exceptions.}\label{tab:appendix_mule_example}
\end{table*}

\begin{table*} [h]
\centering
\small
\resizebox{\textwidth}{!}{
\begin{tabularx}{\textwidth}{X}
    \toprule
    
    \textbf{Question:}\\
    I'm studying ideal and real gasses in engineering thermodynamics. Why can't we measure temperature in Fahrenheit?\\
    \midrule

    \textbf{Predicted answer:}\\
    In engineering thermodynamics, it's common to work with ideal and real gases, and temperature plays a crucial role in characterizing their behavior.\\
    \\
    Actually, you can measure temperature in Fahrenheit, but not in the context of the ideal gas law or ideal gas equations. The issue lies in the definition and scaling of the Fahrenheit and Celsius temperature scales.\\
    \\
    Celsius is the preferred choice for scientific and engineering applications because it's based on the concept of absolute zero, which is the theoretically lowest possible temperature. It's defined as 0 K (kelvin). This allows for a coherent and consistent framework for discussing temperature and thermodynamic quantities.\\
    \\
    Fahrenheit, on the other hand, has a different zero point (32°F) and relies on arbitrary reference points to define its scale. Although it's still used for everyday applications, it doesn't provide the same theoretical foundation as Celsius.\\
    \\
    The ideal gas law and the equations of state for gases are typically expressed in terms of absolute temperature (K or °R), and Celsius is the most compatible scale for these applications.\\
    \\
    So, while you could measure temperature in Fahrenheit if needed, it's essential to understand the underlying science and choose the most relevant and consistent temperature scale when working with ideal and real gases in engineering thermodynamics.\\
    \midrule
    \textbf{Feedback:}\\
    Feedback: The response provides an accurate explanation regarding the use of Fahrenheit versus Celsius (and indirectly Kelvin) in the context of thermodynamics and ideal gases. It correctly identifies the reasons why Celsius and Kelvin are preferred for scientific studies, focusing on the relevance of absolute zero and the uniformity in scientific communication. However, it slightly misleads by stating that Celsius is based on the concept of absolute zero, when in fact, it is Kelvin that is based on this concept. Celsius is indeed more aligned with Kelvin compared to Fahrenheit but it is not based on absolute zero; rather, it is defined by the melting point of ice and the boiling point of water at one atmosphere of pressure, which are then used to extrapolate its relation to Kelvin. The response demonstrates a strong understanding but includes a minor inaccuracy regarding the basis of the Celsius scale. This inaccuracy does not significantly detract from the overall quality and factual correctness of the explanation but is worth noting for precision.\\
    \midrule
    \textbf{Score:}\\
    4\\
    \bottomrule
\end{tabularx}}
\caption{Example LLaMA 8B Instruct zero-shot inference on a $D_3$ question followed by GPT-4 Turbo evaluation.}
\label{tab:appendix_qa_example}
\end{table*}

\begin{example}[h]{Common Backward Discrepancy}[example:appendix_common_bwd]
    \textbf{{\three} question:} I understand the point of the tangle number. But why go to the extra lengths of considering the continued fraction representation of the tangle number.
    
    \vspace{2mm}
    
    \begin{tcolorbox}[
      enhanced jigsaw,
      breakable,
      pad at break*=1mm,
      colback=green!5!white,
      colframe=green!75!black,
      title=LLaMA 2 7B Chat Predicted Answer (Score: 4),
      width=\textwidth
    ]
    (...) \textit{The continued fraction representation of a tangle number allows us to represent the tangle as a rational number}, which can be easily manipulated and analyzed. This can be particularly useful when studying the properties of tangles, such as their convergence properties, their behavior under certain operations, and their relationship to other mathematical structures. (...)
    \end{tcolorbox}
    
    \vspace{2mm}
    
    \begin{tcolorbox}[
      enhanced jigsaw,
      breakable,
      pad at break*=1mm,
      colback=green!5!white,
      colframe=green!75!black,
      title=LLaMA 3 70B Instruct Predicted Answer (Score: 4),
      width=\textwidth
    ]
    (...) By looking at the continued fraction representation, we can gain insight into the tangle's underlying topology and geometry. In particular, the continued fraction representation can reveal information about the tangle's symmetry, its decomposition into simpler components, and even its relationship to other tangles. (...) \end{tcolorbox}
    
    \vspace{2mm}
    
    \textbf{{\two} question:} What is the process of constructing a tangle for a given rational number?
    
    \vspace{2mm}
    
    \begin{tcolorbox}[
      enhanced jigsaw,
      breakable,
      pad at break*=1mm,
      colback=red!5!white,
      colframe=red!75!black,
      title=LLaMA 2 7B Chat Predicted Answer (Score: 1),
      width=\textwidth
    ]
    (...) Choose a base: There are many different bases that you can use to construct a tangle for a rational number. Some common bases include \textit{the golden ratio, the Fibonacci sequence, and the square root of 2}. Choose a base that you feel comfortable working with and that will help you visualize the tangle. (...) \end{tcolorbox}
    
    Feedback: Introduces unrelated concepts that are not part of the accepted method for constructing tangles based on rational numbers.
    
    \vspace{2mm}
    
    \begin{tcolorbox}[
      enhanced jigsaw,
      breakable,
      pad at break*=1mm,
      colback=red!5!white,
      colframe=red!75!black,
      title=LLaMA 3 70B Instruct Predicted Answer (Score: 2),
      width=\textwidth
    ]
    (...) \textit{Draw p arcs}: Draw p arcs from the top of the circle to the bottom, each crossing the circle at a single point. These arcs will form the strands of the tangle.
    
    \textit{Twist the arcs}: Twist the arcs around each other according to the following rule: for each of the q crossings, take the upper strand and pass it over or under the lower strand (it doesn't matter which) to create a twist. (...) 
    
    \end{tcolorbox}
    
    Feedback: Describes a process that diverges from the correct method of using continued fractions to construct a tangle. It also omits the critical step of expressing the rational number as a continued fraction and constructing the tangle diagram based on the continued fraction's terms.
\end{example}

\begin{example}[h]{Backward Discrepancy by Small Model}[example:appendix_small_bwd]
    \textbf{{\three} question:} Given that node operations are O(1), explain whether this implies that all operations in data structures using nodes will also have O(1) complexity, and why or why not.
    
    \vspace{2mm}
    
    \begin{tcolorbox}[
      enhanced jigsaw,
      breakable,
      pad at break*=1mm,
      colback=green!5!white,
      colframe=green!75!black,
      title=LLaMA 2 7B Chat Predicted Answer (Score: 5),
      width=\textwidth
    ]
    (...) \textit{just because node operations have a constant time complexity doesn't necessarily mean that all operations in data structures using nodes will also have O(1) complexity}. The reason for this is that the time complexity of an operation can depend on many factors beyond just the size of the input, such as the data structure's internal organization, the algorithms used, and the hardware on which the data structure is running. (...)
    \end{tcolorbox}
    
    \vspace{2mm}
    
    \textbf{{\two} question:} How does the insertion operation work in a linked list, and what factors affect its time complexity?
    
    \vspace{2mm}
    
    \begin{tcolorbox}[
      enhanced jigsaw,
      breakable,
      pad at break*=1mm,
      colback=red!5!white,
      colframe=red!75!black,
      title=LLaMA 2 7B Chat Predicted Answer (Score: 2),
      width=\textwidth
    ]
    (...) Here are the common ways to perform insertion in a linked list:
    
    1. \textit{Linear Insertion}: (...) This approach has a time complexity of \textit{O(1)} because we only need to traverse the list once to find the correct position.
    2. \textit{Shift-Insertion}: (...) This approach has a time complexity of O(n) because we need to traverse the entire list to shift all the nodes.
    3. \textit{Rotate-Insertion}: (...) This approach has a time complexity of O(n) because we need to traverse the entire list to rotate all the nodes. (...)
    \end{tcolorbox}
    
    Feedback: Introduces non-standard terms (``Linear Insertion,'' ``Shift-Insertion,'' ``Rotate-Insertion'') that are not typically used in describing linked list operations. Incorrectly states that insertion at a specific position has O(1) time complexity, when it should be O(n) in the worst case.
\end{example}

\begin{example}[h]{Backward Discrepancy by Large Model}[example:appendix_large_bwd]
    \textbf{{\three} question:} The quadratic formula is pretty cool. What similar formulas, if any, exist for higher degree equations?
    
    \vspace{2mm}
    
    \begin{tcolorbox}[
      enhanced jigsaw,
      breakable,
      pad at break*=1mm,
      colback=green!5!white,
      colframe=green!75!black,
      title=LLaMA 3 70B Instruct Predicted Answer (Score: 5),
      width=\textwidth
    ]
    (...) For cubic equations of the form $ax^3 + bx^2 + cx + d = 0$, there is a formula known as \textit{Cardano's Formula} (or \textit{Cardano's Method}). (...) For quartic equations of the form $ax^4 + bx^3 + cx^2 + dx + e = 0$, there is a formula known as \textit{Ferrari's Method} (or \textit{Ferrari's Solution}). (...) For equations of degree 5 or higher, there is no general formula that can be written using only addition, subtraction, multiplication, division, and nth roots (like the quadratic formula). This was proven by Niels Henrik Abel in 1824, and later independently by Évariste Galois in 1832. (...)
    \end{tcolorbox}
    
    \vspace{2mm}
    
    \textbf{{\two} question:} Explain the steps to solve quartic equations using the quartic formula.
    
    \vspace{2mm}
    
    \begin{tcolorbox}[
      enhanced jigsaw,
      breakable,
      pad at break*=1mm,
      colback=red!5!white,
      colframe=red!75!black,
      title=LLaMA 3 70B Instruct Predicted Answer (Score: 1),
      width=\textwidth
    ]
    The quartic formula! While it's not as well-known as the quadratic formula, it's a powerful tool for solving quartic equations of the form $ax^4 + bx^3 + cx^2 + dx + e = 0$. Here are the steps to solve quartic equations using the quartic formula:
    
    \textit{Steps to Solve Quartic Equations:}
    
    1. \textit{Write down the quartic equation in standard form}: $ax^4 + bx^3 + cx^2 + dx + e = 0$, where a, b, c, d, and e are constants.\\
    2. \textit{Rearrange the equation to have a = 1}: Divide both sides of the equation by a, so that the coefficient of $x^4$ is 1. This simplifies the formula.\\
    3. \textit{Calculate the values of p, q, and r}: \\
       - $p = b^2 - 3c$\\
       - $q = bc - 4d - a$\\
       - $r = c^2 - 3bd + 12ae$\\
    4. \textit{Calculate the value of u}: $u = (r/2 \pm \sqrt{(r/2)^2 + (q/3)^3})^(1/3)$\\
    5. \textit{Calculate the value of v}: $v = (q + u - b^2/3) / (3u)$\\
    6. \textit{Calculate the value of w}: $w = (-b \pm \sqrt{b^2 - 4v}) / 2$\\
    7. \textit{Solve for x}: $x = (-b \pm 2w \pm \sqrt{w^2 - 4u}) / 2$\\
    \end{tcolorbox}
    
    Feedback: Introduces an incorrect formula for solving quartic equations, which does not align with established methods like Ferrari's solution. The provided steps and calculations do not correspond to the correct procedure for solving quartic equations using the quartic formula or any other recognized method.
\end{example}

\section{Backward Discrepancy Examples}\label{appendix:discrepancy_examples}
Different backward discrepancy examples are shown in Example~\ref{example:appendix_common_bwd}, ~\ref{example:appendix_small_bwd}, and ~\ref{example:appendix_large_bwd}.

\section{Prompts}\label{appendix:prompts}
\subsection{Data construction}\label{appendix:data_prompts}
We provide the prompts used to classify TutorEval questions (Table~\ref{tab:appendix_classify_questions}), generate $D_3$ answers (Table~\ref{tab:appendix_answer_gen_d3}), generate $D_2$ or $D_2$ answers (Table~\ref{tab:appendix_answer_gen}), generate questions at $D_2$ (Table~\ref{tab:appendix_question_gen_d2}) and $D_1$ (Table~\ref{tab:appendix_question_gen_d1}), and augment questions at $D_2$ (Table~\ref{tab:appendix_question_aug_d2}) and $D_1$ (Table~\ref{tab:appendix_question_aug_d1}). For generating or augmenting any question at $D_2$ or $D_1$, we use the same system prompt (Table~\ref{tab:appendix_question_gen_system}) that describes the definitions of depths of knowledge. 

\subsection{Inference} \label{appendix:inference_prompts}
We provide the prompts used for zero-shot (Table~\ref{tab:appendix_zeroshot}), Prompt (Gold) and Prompt (Pred.) (Table~\ref{tab:appendix_ctx}), and multi-turn (Table~\ref{tab:appendix_multiturn}) inference.



\subsection{Evaluation}\label{appendix:evaluation_prompts}
The prompt used for LLM-as-a-Judge evaluation is in Table~\ref{tab:appendix_eval}.

\begin{table*} [h]
\centering
\small
\resizebox{\textwidth}{!}{
\begin{tabularx}{\textwidth}{X}
    \toprule
    \textbf{System prompt:}\\
    You are an excellent question classifier. You will be given (1) a question and (2) key points that a good response would address when answering the question. You have to classify at which Depth of Knowledge (DOK) level the question is located. DOK is a framework that focuses on the context which knowledge will be demonstrated. Here is the definition of each DOK level:\\
    \\
    1. DOK-1 (Basic Knowledge and Recall): This level addresses ``What is the knowledge?''. It evaluates the ability to remember, explain, or pinpoint fundamental facts, terms, principles, and procedures. It's about recognizing or recollecting basic information and performing simple, direct tasks.\\
    2. DOK-2 (Application of Knowledge and Skills): This level explores ``How can the knowledge be used?''. It tests the ability to employ knowledge and concepts in practical situations, which involves choosing appropriate methods, solving straightforward problems, or interpreting data. This level acts as an intermediary step between fundamental understanding and more advanced reasoning.\\
    3. DOK-3 (Analytical and Strategic Thinking): This level questions ``Why can the knowledge be used?''. It challenges one to use strategic thought, logic, and problem-solving in intricate, abstract situations that might have more than one solution. This stage demands critical thinking, rationale, and conceptualization of theoretical scenarios.\\
    4. DOK-4 (Extended and Integrative Knowledge): This level examines ``How else can the knowledge be applied?''. It assesses the ability to conduct thorough research, apply concepts and skills in real-world scenarios, and integrate knowledge across different disciplines or sources. It involves developing original ideas, conducting experiments, and synthesizing information from various fields. Note that in the science domain, this level may be constrained to designing studies, experiments, and projects and is thus rare or even absent in most standardized assessment.\\

    \midrule
    
    \textbf{User prompt:}\\
    Please classify the following question into DOK-1, 2, 3, or 4. Refer to the key points to help your judgment. Think step-by-step and provide an explanation of your judgment. After providing your explanation, output the DOK level that is an integer of 1, 2, 3, or 4. The output format should looks as follows: \{explanation for reaching the DOK decision\} \lbrack RESULT \rbrack \{DOK level that is an integer in the range 1 to 4\}.\\
    \\
    \#\# Question\\
    \textcolor{violet}{\texttt{\{question\}}}\\
    \#\# Key points\\
    \textcolor{violet}{\texttt{\{key\_points\}}}\\
    \#\# Answer\\
    \bottomrule
\end{tabularx}}
\caption{Prompt for classifying TutorEval questions.}
\label{tab:appendix_classify_questions}
\end{table*}

\begin{table*}[h]
\centering
\small
\resizebox{\textwidth}{!}{
\begin{tabularx}{\textwidth}{X}
    \toprule
    \textbf{System prompt:}\\
    You are an excellent assistant that effectively answers complex questions. You are given a passage, question, and key points to answer the question. Read the instruction and give an appropriate answer.\\
    \midrule
    
    \textbf{User prompt:}\\
    \#\# Chapter\\
    \textcolor{violet}{\texttt{\{chapter\}}}\\
    \\
    \#\# Instruction\\
    Answer the question below. \\
    - You may refer to the contents in the chapter text above if necessary, but do NOT expose in your answer that you are referring to the provided source. \\
    - Ensure that the answer is complete, fully satisfying the key points to answer the question. \\
    - The answer must also match the level of cognitive complexity required, incorporating the context which the depth of knowledge will be demonstrated. \\
    \\
    \#\# Question\\
    \textcolor{violet}{\texttt{\{question\}}}\\
    \\
    \#\# Key points to answer the question\\
    \textcolor{violet}{\texttt{\{key\_points\}}}\\
    \\
    \#\# Complexity of the question\\
    \textcolor{violet}{\texttt{\{explanation\}}}\\
    \\
    \#\# Answer\\
    \bottomrule
\end{tabularx}}
\caption{Prompt for generating reference answer for a $D_3$ question.}
\label{tab:appendix_answer_gen_d3}
\end{table*}

\begin{table*}[h]
\centering
\small
\resizebox{\textwidth}{!}{
\begin{tabularx}{\textwidth}{X}
    \toprule
    \textbf{System prompt:}\\
    You are a helpful assistant that accurately answers complex questions. Ensure that your answer is focused and compact.\\
    \midrule
    
    \textbf{User prompt:}\\
    \textcolor{violet}{\texttt{\{question\}}}\\
    \bottomrule
\end{tabularx}}
\caption{Prompt for generating reference answer for a $D_1$ or $D_2$ question.}
\label{tab:appendix_answer_gen}
\end{table*}

\begin{table*}[h]
\centering
\small
\resizebox{\textwidth}{!}{
\begin{tabularx}{\textwidth}{X}
    \toprule
    \textbf{System prompt:}\\
    You are an excellent question generator. You will be given a question and a gold answer to the question. You have to generate shallower questions from the given question. Here is the definition of the depth of knowledge a question requires:\\
    \\
    1. Depth-1 (Basic Knowledge and Recall): This level addresses ``What is the knowledge?''. It evaluates the ability to remember, explain, or pinpoint fundamental facts, terms, principles, and procedures. It's about recognizing or recollecting basic information and performing simple, direct tasks.\\
    2. Depth-2 (Application of Knowledge and Skills): This level explores ``How can the knowledge be used?''. It tests the ability to employ knowledge and concepts in practical situations, which involves choosing appropriate methods, solving straightforward problems, or interpreting data. This level acts as an intermediary step between fundamental understanding and more advanced reasoning.\\
    3. Depth-3 (Analytical and Strategic Thinking): This level questions ``Why can the knowledge be used?''. It challenges one to use strategic thought, logic, and problem-solving in intricate, abstract situations that might have more than one solution. This stage demands critical thinking, rationale, and conceptualization of theoretical scenarios.\\
    \bottomrule
\end{tabularx}}
\caption{System prompt for generating or augmenting $D_1$ or $D_2$ questions.}
\label{tab:appendix_question_gen_system}
\end{table*}

\begin{table*}[h]
\centering
\small
\resizebox{\textwidth}{!}{
\begin{tabularx}{\textwidth}{X}
    \toprule
    \textbf{User prompt:}\\
    \#\# Instruction\\
    Create maximum of 4 Depth-2 questions that are necessary to answer the provided Depth-3 question correctly. \\
    - Remember that Depth-2 questions are centered on application of procedural knowledge and skills and Depth-3 questions are centered on analysis and strategic knowledge.\\
    - Take into consideration the level of cognitive complexity required to solve the Depth-3 question, so that your generated questions fall under the description of Depth-2 appropriately.\\
    - Ensure that your collection of generated Depth-2 questions adequately and comprehensively covers ALL the necessary factual or conceptual knowledge required to answer the given Depth-3 question.\\
    - Ensure that all of your generated Depth-2 questions do not directly answer to the given Depth-3 question.\\
    - The number of generated Depth-2 questions should not exceed 4.\\
    - The generated Depth-2 questions should be in JSON format: \{``Depth-2\_questions'': \lbrack list of Depth-2 question strings \rbrack\}\\
    \\
    \#\# Example 1\\
    \#\#\# Depth-3 question\\
    What is the intuition behind the Gram - Schmidt procedure?\\
    \#\#\# Generated Depth-2 questions\\
    \{``Depth-2\_questions'': \lbrack 'How do you project one vector onto another vector?', 'What does it mean for two vectors to be orthogonal, and how can you verify this property?', 'Describe the process of normalizing a vector.', 'Explain how subtracting the projection of one vector from another results in orthogonality.', 'Given a set of vectors, how can you determine if they are linearly independent?', 'How can the concept of linear independence be used to form a basis for a vector space?' \rbrack\}\\
    \\
    \#\# Example 2\\
    \#\#\# Depth-3 question\\
    Why couldn't we test general relativity effects using the Eotvos experiment?\\
    \#\#\# Generated Depth-2 questions\\
    \{``Depth-2\_questions'': \lbrack ``How does the Eötvös experiment determine the equivalence between inertial mass and gravitational mass?'', ``Describe the Equivalence Principle and its significance in the theory of General Relativity.'', ``Identify experiments or observations that could directly test the predictions of General Relativity, such as time dilation or the bending of light.'', ``How do experiments measuring time dilation differ in design and scope from those measuring mass equivalence?'' \rbrack\}\\
    \\
    \#\# Example 3\\
    \#\#\# Depth-3 question\\
    Why are aldehydes more readily oxidized to carboxylic acids compared to ketones, and how does this difference in reactivity influence their identification in the laboratory?\\
    \#\#\# Generated Depth-2 questions\\
    \{``Depth-2\_questions'': \lbrack ``How can you identify an aldehyde using Tollens' reagent?'', ``Why does the carbonyl carbon in aldehydes have a significant partial positive charge?'', ``How does the structure of ketones differ from that of aldehydes, and how does this affect their reactivity towards oxidation?'' \rbrack\}\\
    \\
    \#\# Example 4\\
    \#\#\# Depth-3 question\\
    In the context of computer programming, why is branching unstructured? And is it a bad design choice?\\
    \#\#\# Generated Depth-2 questions\\
    \{``Depth-2\_questions'': \lbrack ``What are the key differences between structured and unstructured branching in programming?'', ``How does the 'goto' statement work in computer programming?'', ``What are the potential risks involved with using unstructured branching in large software projects?'', ``How does the structure of a program affect its maintainability?'', ``How can the flow of execution in a program influence its debuggability?'' \rbrack\}\\
    \\
    \#\# Depth-3 question\\
    \textcolor{violet}{\texttt{\{question\}}}\\
    \\
    \#\# Answer to the Depth-3 question\\
    \textcolor{violet}{\texttt{\{answer\}}}\\
    \\
    \#\# Generated Depth-2 questions\\
    \bottomrule
\end{tabularx}}
\caption{User prompt for generating $D_2$ questions.}
\label{tab:appendix_question_gen_d2}
\end{table*}

\begin{table*}[h]
\centering
\small
\resizebox{\textwidth}{!}{
\begin{tabularx}{\textwidth}{X}
    \toprule
    \textbf{User prompt:}\\
    \#\# Instruction\\
    Create maximum of 4 Depth-1 questions that are necessary to answer the provided Depth-2 question correctly. \\
    - Remember that Depth-1 questions are centered on basic recall of factual and conceptual knowledge. Depth-2 questions are centered on application of procedural knowledge and skills.\\
    - Take into consideration the level of cognitive complexity required to solve the Depth-2 question, so that your generated questions fall under the description of Depth-1 appropriately.\\
    - Ensure that your collection of generated Depth-1 questions adequately and comprehensively covers ALL the necessary factual or conceptual knowledge required to answer the given Depth-2 question.\\
    - Ensure that all of your generated Depth-1 questions do not directly answer to the given Depth-2 question.\\
    - Try to exclude Depth-1 questions that ask too generic or commonsense knowledge.\\
    - The number of generated DOK-2 questions should not exceed 4.\\
    - The generated Depth-1 questions should be in JSON format: \{``Depth-1\_questions'': \lbrack list of Depth-1 question strings\rbrack\}\\
    \\
    \#\# Example 1\\
    \#\#\# Depth-2 question\\
    How can the concept of algebraic closure be demonstrated using polynomial equations with complex roots?\\
    \#\#\# Generated Depth-1 questions\\
    \{``Depth-1\_questions'': \lbrack 'What is the definition of algebraic closure?', 'What is a polynomial equation?', 'What are complex roots in the context of polynomial equations?', 'How can complex roots be represented?'\rbrack\}\\
    \\
    \#\# Example 2\\
    \#\#\# Depth-2 question\\
    How do you perform a convolution operation between two random variables?\\
    \#\#\# Generated Depth-1 questions\\
    \{``Depth-1\_questions'': \lbrack 'What is a convolution operation?', 'What is a random variable?', 'How is the product of two functions calculated?', 'What does it mean to integrate a function?'\rbrack\}\\
    \\
    \#\# Example 3\\
    \#\#\# Depth-2 question\\
    In what ways can a decision tree's structure be represented programmatically?\\
    \#\#\# Generated Depth-1 questions\\
    \{``Depth-1\_questions'': \lbrack 'What is a decision tree in the context of programming?', 'What are the basic components of a decision tree?', 'What is a data structure in programming?', ``What does 'represented programmatically' mean?''\rbrack\}\\
    \\
    \#\# Example 4\\
    \#\#\# Depth-2 question\\
    How do neutrinos differ from other subatomic particles, and why are they considered potential candidates for dark matter?\\
    \#\#\# Generated Depth-1 questions\\
    \{``Depth-1\_questions'': \lbrack 'What are neutrinos?', 'What are subatomic particles?', 'What is dark matter?', 'What characteristics do particles need to be considered candidates for dark matter?'\rbrack\}\\
    \\
    \#\# Depth-2 question\\
    \textcolor{violet}{\texttt{\{question\}}}\\
    \\
    \#\# Answer to the Depth-2 question\\
    \textcolor{violet}{\texttt{\{answer\}}}\\
    \\
    \#\# Generated Depth-1 questions\\
    \bottomrule
\end{tabularx}}
\caption{User prompt for generating $D_1$ questions.}
\label{tab:appendix_question_gen_d1}
\end{table*}

\begin{table*}[h]
\centering
\small
\resizebox{\textwidth}{!}{
\begin{tabularx}{\textwidth}{X}
    \toprule
    \textbf{User prompt:}\\
    \#\# Instruction\\
    Create \textcolor{violet}{\texttt{\{count\}}} Depth-2 question(s) that complement current Depth-2 questions, which are necessary to correctly answer the provided Depth-3 question. \\
    - Remember that Depth-2 questions are centered on application of procedural knowledge and skills and Depth-3 questions are centered on analysis and strategic knowledge.\\
    - Take into consideration the level of cognitive complexity required to solve the Depth-3 question, so that your generated questions fall under the description of Depth-2 appropriately.\\
    - Complement the existing Depth-2 questions with additional questions to ensure they collectively cover all necessary procedural knowledge and skills required to answer the Depth-3 question effectively.\\
    - Ensure that all of your generated Depth-2 questions do not directly answer to the given Depth-3 question.\\
    - The number of all Depth-2 questions should not exceed 4.\\
    - The generated Depth-2 questions should be in JSON format: \{``Depth-2\_questions'': \lbrack list of Depth-2 question strings \rbrack \}\\
    \\
    \#\# Example 1\\
    \#\#\# Depth-3 question and current Depth-2 questions\\
    What is the intuition behind the Gram - Schmidt procedure?\\
    \{``current\_Depth-2\_questions'': \lbrack 'How do you project one vector onto another vector?', 'What does it mean for two vectors to be orthogonal, and how can you verify this property?', 'Describe the process of normalizing a vector.', 'Explain how subtracting the projection of one vector from another results in orthogonality.', 'Given a set of vectors, how can you determine if they are linearly independent?' \rbrack \}\\
    \#\#\# Generated complementary Depth-2 questions\\
    \{``complementary\_Depth-2\_questions'': \lbrack 'How can the concept of linear independence be used to form a basis for a vector space?' \rbrack \}\\
    \\
    \#\# Example 2\\
    \#\#\# Depth-3 question and current Depth-2 questions\\
    Why couldn't we test general relativity effects using the Eotvos experiment?\\
    \{``current\_Depth-2\_questions'': \lbrack ``How does the Eötvös experiment determine the equivalence between inertial mass and gravitational mass?'', ``Describe the Equivalence Principle and its significance in the theory of General Relativity.'', ``Identify experiments or observations that could directly test the predictions of General Relativity, such as time dilation or the bending of light.'' \rbrack \}\\
    \#\#\# Generated complementary Depth-2 questions\\
    \{``complementary\_Depth-2\_questions'': \lbrack ``How do experiments measuring time dilation differ in design and scope from those measuring mass equivalence?'' \rbrack \}\\
    \\
    \#\# Example 3\\
    \#\#\# Depth-3 question and current Depth-2 questions\\
    Why are aldehydes more readily oxidized to carboxylic acids compared to ketones, and how does this difference in reactivity influence their identification in the laboratory?\\
    \{``current\_Depth-2\_questions'': \lbrack ``How can you identify an aldehyde using Tollens' reagent?'', ``Why does the carbonyl carbon in aldehydes have a significant partial positive charge?'' \rbrack \}\\
    \#\#\# Generated complementary Depth-2 questions\\
    \{``complementary\_Depth-2\_questions'': \lbrack ``How does the structure of ketones differ from that of aldehydes, and how does this affect their reactivity towards oxidation?'' \rbrack \}\\
    \\
    \#\# Example 4\\
    \#\#\# Depth-3 question and current Depth-2 questions\\
    In the context of computer programming, why is branching unstructured? And is it a bad design choice?\\
    \{``current\_Depth-2\_questions'': \lbrack ``What are the key differences between structured and unstructured branching in programming?'', ``How does the 'goto' statement work in computer programming?'' \rbrack \}\\
    \#\#\# Generated complementary Depth-2 questions\\
    \{``complementary\_Depth-2\_questions'': \lbrack ``What are the potential risks involved with using unstructured branching in large software projects?'', ``How does the structure of a program affect its maintainability?'', ``How can the flow of execution in a program influence its debuggability?'' \rbrack \}\\
    \\
    \#\# Depth-3 question\\
    \textcolor{violet}{\texttt{\{question\}}}\\
    \\
    \#\# Answer to the Depth-3 question\\
    \textcolor{violet}{\texttt{\{answer\}}}\\
    \\
    \#\# Current Depth-2 questions\\
    \{``current\_Depth-2\_questions'': \textcolor{violet}{\texttt{\{current\_questions\}}}\}\\
    \\
    \#\# Generated \textcolor{violet}{\texttt{\{count\}}} complementary Depth-2 questions\\
    \bottomrule
\end{tabularx}}
\caption{User prompt for augmenting $D_2$ questions.}
\label{tab:appendix_question_aug_d2}
\end{table*}

\begin{table*}[h]
\centering
\small
\resizebox{\textwidth}{!}{
\begin{tabularx}{\textwidth}{X}
    \toprule
    \textbf{User prompt:}\\
    \#\# Instruction\\
    Create \textcolor{violet}{\texttt{\{count\}}} Depth-1 question(s) that complement current Depth-1 questions, which are necessary to correctly answer the provided Depth-2 question. \\
    - Remember that Depth-1 questions are centered on basic recall of factual and conceptual knowledge. Depth-2 questions are centered on application of procedural knowledge and skills.\\
    - Take into consideration the level of cognitive complexity required to solve the Depth-2 question, so that your generated questions fall under the description of Depth-1 appropriately.\\
    - Complement the existing Depth-1 questions with additional questions to ensure they collectively cover all necessary procedural knowledge and skills required to answer the Depth-2 question effectively.\\
    - Ensure that all of your generated Depth-1 questions do not directly answer to the given Depth-2 question.\\
    - Try to exclude Depth-1 questions that ask too generic or commonsense knowledge.\\
    - The number of all Depth-1 questions should not exceed 4.\\
    - The generated Depth-1 questions should be in JSON format: \{``complementary\_Depth-1\_questions'': \lbrack list of Depth-1 question strings\rbrack\}\\
    \\
    \#\# Example 1\\
    \#\#\# Depth-2 question and current Depth-1 questions\\
    How can the concept of algebraic closure be demonstrated using polynomial equations with complex roots?\\
    \{``current\_Depth-1\_questions'': \lbrack'What is the definition of algebraic closure?', 'What is a polynomial equation?', 'What are complex roots in the context of polynomial equations?'\rbrack\}\\
    \#\#\# Generated complementary Depth-1 questions\\
    \{``complementary\_Depth-1\_questions'': \lbrack'How can complex roots be represented?'\rbrack\}\\
    \\
    \#\# Example 2\\
    \#\#\# Depth-2 question and current Depth-1 questions\\
    How do you perform a convolution operation between two random variables?\\
    \{``current\_Depth-1\_questions'': \lbrack'What is a convolution operation?', 'What is a random variable?', 'How is the product of two functions calculated?'\rbrack\}\\
    \#\#\# Generated complementary Depth-1 questions\\
    \{``complementary\_Depth-1\_questions'': \lbrack'What does it mean to integrate a function?'\rbrack\}\\
    \\
    \#\# Example 3\\
    \#\#\# Depth-2 question and current Depth-1 questions\\
    In what ways can a decision tree's structure be represented programmatically?\\
    \{``current\_Depth-1\_questions'': \lbrack'What is a decision tree in the context of programming?', 'What are the basic components of a decision tree?'\rbrack\}\\
    \#\#\# Generated complementary Depth-1 questions\\
    \{``complementary\_Depth-1\_questions'': \lbrack'What is a data structure in programming?', ``What does 'represented programmatically' mean?''\rbrack\}\\
    \\
    \#\# Example 4\\
    \#\#\# Depth-2 question and current Depth-1 questions\\
    How do neutrinos differ from other subatomic particles, and why are they considered potential candidates for dark matter?\\
    \{``current\_Depth-1\_questions'': \lbrack'What are neutrinos?', 'What are subatomic particles?'\rbrack\}\\
    \#\#\# Generated complementary Depth-1 questions\\
    \{``complementary\_Depth-1\_questions'': \lbrack'What is dark matter?', 'What characteristics do particles need to be considered candidates for dark matter?'\rbrack\}\\
    \\
    \#\# Depth-2 question\\
    \textcolor{violet}{\texttt{\{question\}}}\\
    \\
    \#\# Answer to the Depth-2 question\\
    \textcolor{violet}{\texttt{\{answer\}}}\\\\
    \\
    \#\# Current Depth-1 questions\\
    \{``current\_Depth-1\_questions'': \textcolor{violet}{\texttt{\{current\_questions\}}}\}\\
    \\
    \#\# Generated \textcolor{violet}{\texttt{\{count\}}} complementary Depth-1 questions\\
    \bottomrule
\end{tabularx}}
\caption{User prompt for augmenting $D_1$ questions.}
\label{tab:appendix_question_aug_d1}
\end{table*}

\begin{table*}[h]
\centering
\small
\begin{tabularx}{0.6\textwidth}{X}
    \toprule
    \textbf{System prompt:}\\
    You are a helpful, respectful and honest assistant. Answer the question.\\
    \midrule
    \textbf{User prompt:}\\
    \#\# Question: \\
    \textcolor{violet}{\texttt{\{$D_{k}$ question\}}}\\
    \\
    \#\# Answer: \\
    \bottomrule
\end{tabularx}
\caption{Prompt for zero-shot inference.}
\label{tab:appendix_zeroshot}
\end{table*}

\begin{table*}[h]
\centering
\small
\begin{tabularx}{0.6\textwidth}{X}
    \toprule
    \textbf{System prompt:}\\
    You are a helpful, respectful and honest assistant. Answer the question.\\
    \midrule
    \textbf{User prompt:}\\
    \#\# QA pairs: \\
    Q: \textcolor{violet}{\texttt{\{$D_{k-1}$ question 1\}}}\\
    A: \textcolor{violet}{\texttt{\{$D_{k-1}$ answer 1\}}}\\
    Q: \textcolor{violet}{\texttt{\{$D_{k-1}$ question 2\}}}\\
    A: \textcolor{violet}{\texttt{\{$D_{k-1}$ answer 2\}}}\\
    \textcolor{violet}{\texttt{...}}
    \\
    \#\# Question: \\
    \textcolor{violet}{\texttt{\{$D_{k}$ question\}}}\\
    \\
    \#\# Answer: \\
    \bottomrule
\end{tabularx}
\caption{Prompt for inference given reference answers or self-predictions on shallower questions.}
\label{tab:appendix_ctx}
\end{table*}

\begin{table*}[h]
\centering
\small
\begin{tabularx}{0.6\textwidth}{X}
    \toprule
    \textbf{System prompt:}\\
    You are a helpful, respectful and honest assistant. Answer the question.\\
    \midrule
    \textbf{User prompt:}\\
    \#\# Question: \\
    \textcolor{violet}{\texttt{\{$D_{k-1}$ question\}}}\\
    \\
    \#\# Answer: \\
    \midrule
    \textbf{[Last turn] System prompt:}\\
    You are a helpful, respectful and honest assistant. Answer the question.\\
    \midrule
    \textbf{User prompt:}\\
    Based on previous questions, answer the question. 
    \#\# Question: \\
    \textcolor{violet}{\texttt{\{$D_{k}$ question\}}}\\
    \\
    \#\# Answer: \\
    \bottomrule
\end{tabularx}
\caption{Prompt for multi-turn inference.}
\label{tab:appendix_multiturn}
\end{table*}

\begin{table*}[h]
\centering
\small
\resizebox{\textwidth}{!}{
\begin{tabularx}{\textwidth}{X}
    \toprule
    \textbf{System prompt:}\\
    You are a fair judge assistant tasked with providing clear, objective feedback based on specific criteria, ensuring each assessment reflects the absolute standards set for performance.\\
    \midrule
    \textbf{User prompt:}\\
    \#\#\#Task Description:\\
    An instruction (might include an Input inside it), a response to evaluate, and a score rubric representing a evaluation criteria are given.\\
    1. Write a detailed feedback that assess the quality of the response strictly based on the given score rubric, not evaluating in general.\\
    2. After writing a feedback, write a score that is an integer between 1 and 5. You should refer to the score rubric.\\
    3. The output format should look as follows: ``Feedback: (write a feedback for criteria) \lbrack RESULT \rbrack (an integer number between 1 and 5)''\\
    4. Please do not generate any other opening, closing, and explanations.\\
    \\
    \#\#\#The instruction to evaluate:\\
    \textcolor{violet}{\texttt{\{instruction\}}}\\
    \\
    \#\#\#Response to evaluate:\\
    \textcolor{violet}{\texttt{\{response\}}}\\
    \\
    \#\#\#Reference Answer (Score 5):\\
    \textcolor{violet}{\texttt{\{reference\_answer\}}}\\
    \\
    \#\#\#Score Rubrics:\\
    \lbrack Is the response correct, accurate, and factual? \rbrack\\
    Score 1: The response is largely incorrect, inaccurate, and not factual. It demonstrates a fundamental misunderstanding of the query or topic, leading to irrelevant or completely erroneous information.\\
    Score 2: The response is partially correct but contains significant inaccuracies or factual errors. It shows some understanding of the query or topic but fails to provide a fully accurate or reliable answer.\\
    Score 3: The response is generally correct and factual but may include minor inaccuracies or lack of detail. It shows a good understanding of the query or topic but may miss some nuances or specific information.\\
    Score 4: The response is mostly correct, accurate, and factual. It demonstrates a strong understanding of the query or topic, with only minimal inaccuracies or omissions that do not significantly detract from the overall quality of the response.\\
    Score 5: The response is consistently correct, accurate, and entirely factual. It reflects a comprehensive understanding of the query or topic, providing detailed, precise, and fully reliable information without any inaccuracies or omissions.\\
    \\
    \#\#\#Feedback: \\
    \bottomrule
\end{tabularx}}
\caption{Prompt for LLM-as-a-Judge evaluation with an accuracy score rubric.}
\label{tab:appendix_eval}
\end{table*}

\end{document}